# A Novel Fusion of Sentinel-1 and Sentinel-2 with Climate Data for Crop Phenology Estimation using Machine Learning

Shahab Aldin Shojaeezadeh[1*], Abdelrazek Elnashar[1, 2], Tobias Karl David Weber[1]
1 Section of Soil Science, Faculty of Organic Agricultural Sciences, University of Kassel, Witzenhausen 37213, Germany
2 Department of Natural Resources, Faculty of African Postgraduate Studies, Cairo University, Giza 12613, Egypt
*Corresponding Author: shahab@uni-kassel.de

## Abstract

Crop phenology describes the physiological development stages of crops from planting to harvest which is valuable information for decision makers to plan and adapt agricultural management strategies. In the era of big Earth observation data ubiquity, attempts have been made to accurately detect crop phenology using Remote Sensing (RS) and high resolution weather data. However, most studies have focused on large scale predictions of phenology or developed methods which are not adequate to help crop modeler communities on leveraging Sentinel-1 and Sentinal-2 data and fusing them with high resolution climate data, using a novel framework. For this, we trained a Machine Learning (ML) LightGBM model to predict 13 phenological stages for eight major crops across Germany at 20 m scale. Observed phonologies were taken from German national phenology network (German Meteorological Service; DWD) between 2017 and 2021. We proposed a thorough feature selection analysis to find the best combination of RS and climate data to detect phenological stages. At national scale, predicted phenology resulted in a reasonable precision of $R^2 > 0.43$ and a low Mean Absolute Error of 6 days, averaged over all phenological stages and crops. The spatio-temporal analysis of the model predictions demonstrates its transferability across different spatial and temporal context of Germany. The results indicated that combining radar sensors with climate data yields a very promising performance for a multitude of practical applications. Moreover, these improvements are expected to be useful to generate highly valuable input for crop model calibrations and evaluations, facilitate informed agricultural decisions, and contribute to sustainable food production to address the increasing global food demand.

**Keywords**: Crop monitoring, Remote sensing, Radar data, Optical data, Google Earth Engine

## 1. Introduction

The phenological development stages determine the onset and duration of plant growth events. Knowledge on these phenological stages plays an important role in agricultural practices in guiding decision makers to plan irrigation schedules and fertilization strategies (Meroni et al. 2021) at local scales. It is also important for monitoring plant productivity, plant health, and identify the incidence of pests and diseases (Xia et al. 2015). Phenological development patterns in natural landscapes could also serve as an indicator for biodiversity (Viña et al. 2016) that enables the evaluation of the impact of climate change (Badeck et al. 2004) and land-use alterations on ecosystems (Morellato et al. 2016). Thus, applications for accurate and precise knowledge on the state of phenology on large spatial scales at high spatial resolution exist, yet the efforts to estimate crop phenological stages accurately are still an ongoing challenge.

Crop phenology is typically assessed by laborious in-situ field observations usually limited to point locations within confined regions. To overcome the resulting data scarcity, researchers have proposed various methods to estimate crop phenological events at large spatial scales based on climate data (Gerstmann et al. 2016; Li et al. 2021) and Remote Sensing (RS) data (Babcock et al. 2021; Tian et al. 2021; Vijaywargiya and Nidamanuri 2023). RS data, particularly multispectral imagery such as Sentinel-2 (S2) optical data, has been used to estimate phenology at regional to global scales (Katal et al. 2022; Tran et al. 2023; Yang et al. 2023a; Yue et al. 2025), and Sentinel-1 (S1) Synthetic Aperture Radar (SAR) data is gaining attention because of its weather resilience (Li et al. 2023; Wang et al. 2019b; Zhao et al. 2022). While radar data have been shown to provide valuable information about phenological developments of winter wheat (Lobert et al. 2023; Mimić et al. 2025; Schlund and Erasmi 2020), rice (Lopez-Sanchez et al. 2011; Lopez-Sanchez et al. 2013), maize (Htitiou et al. 2024), sugar beet (Htitiou et al. 2024; Löw et al. 2021), and other crops (d'Andrimont et al. 2020; Meroni et al. 2021; Wang et al. 2019a), the performance of these approaches are still under debate (Mercier et al. 2020; Meroni et al. 2021); underlining the need for further exploration. In addition, various studies showed that radar and multispectral data fusion is ideal to estimate phenological development (De Bernardis et al. 2016; Lobert et al. 2023; Mercier et al. 2020; Meroni et al. 2021). Furthermore, research has suggested to blend climatic data with both radar and optical data (Nieto et al. 2021; Zhou et al. 2024). Conversely, another side contends that incorporating climate features into a satellite data driven approach, does not help to estimate the onset of crop growing stages (Lobert et al. 2023). Hence, fusing optical and radar imageries along with climate data on detecting plant growth stages needs further assessment to bridge the gap between field, experimental studies, and RS and climate data.

Various methods using RS data have been developed to estimate phenological development stages. These include analyzing time series of Vegetation Indices (VIs), well-known as Land Surface Phenology (LSP, (Babcock et al. 2021; Nietupski et al. 2021; Tran et al. 2023; Tran et al. 2025; Yang et al. 2023a)), combining VIs and/or satellite bands (i.e., raw band data) with Physical Crop Models (PCM, (MacBean et al. 2015; Viswanathan et al. 2022; Worrall et al. 2023)), Machine Learning algorithms (ML, (Katal et al. 2022; Li et al. 2021; Lobert et al. 2023; Ma et al. 2023; Zhou et al. 2021)), and phenology matching models such as shape model fitting (SMF, (Diao et al. 2021; Liu et al. 2022)) at various regional and global scales (Tran et al. 2023). LSP is a well-known method that focusses on overall crop growth stages to estimate phenological events at the start (SoS) and end (EoS) of the growing season; for more information refer to (Zeng et al. 2020). While, recent studies focused on methods that explicitly estimate predefined phenological stages coinciding with ground observations (Canisius et al. 2018; Diao et al. 2021; Liu et al. 2022; Lobert et al. 2023; Wang et al. 2019b), there is still a need to enhance the proposed methods based on RS data. With today's richness of RS and climate data and advancements in ML models, there is a lot of hope to synergize ML-RS-Climate-based model imputed with ground observations to improve the accuracy of estimating phenological stages of crops (Kooistra et al. 2023). Although various studies explored the potential of ML models to predict phenology (Czernecki et al. 2018; Wang et al. 2023; Worrall et al. 2023; Xin et al. 2020; Yang et al. 2023b), these studies are often limited to specific crops or phenological stages, limiting their applicability to a broader range of agricultural scenarios (Lobert et al. 2023; Tedesco et al. 2021) and limiting their incorporation into decision support algorithms for agriculture.

Therefore, this study aims to detect crop phenology stages by fusing S1 and S2 data, and climate data along with a Crop Type Map (CTM; Blickensdörfer et al. (2022)) of Germany. To this end, we analyzed the contribution of various vegetation indices, climate parameters, and static features such as elevation and geolocation as input features for a Tree-based gradient boosting ML algorithm of Light Gradient-Boosting Machine (LightGBM). As ground truth, the database of the German Weather Service on phenological observation covering all of Germany was used. The study focused on eight major crops – maize, spring and winter barley, spring oat, sugar beet, winter rapeseed, winter rye, and winter wheat – and encompassed in total 13 phenological stages. The study period covered the years 2017 through 2021. Specifically, the following questions were addressed:

1- Which combinations of RS and climate data perform best in our novel ML based fusion method to predict phenological stages?
2- What factors influence the ML model's performance across different phenological stages and crops?
3- How effectively do RS and climate data represent the spatio-temporal variations of phenological events across Germany, and how transferable are these insights over different regions and time periods?

# 2. Materials and methods

## 2.1. Study area and phenology data

The comprehensive phenology database of the German Meteorological Service (DWD) was used as reference data. This database is well-known for its long-term volunteer-based observations (~1200 trained observers) that started in 1950 with over 10 million observations and 1000 stations across Germany (Kaspar et al. 2015). However, these observations are not geotagged to specific locations; instead, they were collected from areas near the stations (see Section 2.2). The DWD phenology database includes phenological stages for various crop types. Maize (corn; *Zea mays L.*), spring and winter barley (*Hordeum vulgare L.*), spring oat (*Avena sativa L.*), sugar beet (*Beta vulgaris L.*), winter rapeseed (*Brassica napus L.*), winter rye (*Secale cereale L.*), and winter wheat (*Triticum aestivum L.*) were selected among others that had consistent observations through time. We chose 862 stations across Germany including more than 86600 observations covering thirteen specific phenological stages between 2017 and 2021. The study area and DWD stations are shown in Fig. 1A.

The DWD has an explicit definition for phenology stages; however, Phenological Development Stages of Plants (BBCH) standard was used in this study to be in line with larger scientific community and for operational goals (Kaspar et al. 2015). The phenological stages that are available include seeding (BBCH~0), emergence or lead development (~10), rosette formation (~14), growth in height or shooting or stem elongation (~31), closed stand (~35), bud formation, heading or tassel emergence (~51), the tip of tassel visible (~53), the beginning of flowering (~61), general flowering (~65), milk ripeness (~75), wax-ripe stage (~83), full or yellow ripeness (~87), and harvesting (~89). Note that different crops have different sets of reported phenological stages such that not all of the crop specific phenology data are covered by all 13 stages (Table 1). For the harvest stage, the exact BBCH scale is not available, but we selected BBCH equal to 89 before starting the latest phase of the plant (beginning of dormancy or senescence~90). Each record in the database holds a quality flag that we used to select only the observations that had no objections during postprocessing and quality control completed (QB=1; no objection | QN=10; quality control finished, all corrections finished) (Kaspar et al. 2015). Furthermore, we removed observations that do not follow the order of BBCH stages based on the date in each station. With this procedure, we are sure that our data is less affected by errors and the weekend bias (Courter et al. 2013). The number of BBCH observations per crop and the number of observations is shown Table 1.

**Table 1** Details on the available BBCH observations and number of data points per crop .

| Crop/BBCH | 0 | 10 | 14 | 31 | 35 | 51 | 53 | 61 | 65 | 75 | 83 | 87 | 89 | Total |
|---|---|---|---|---|---|---|---|---|---|---|---|---|---|---|
| Maize | 2525 | 2590 | | 2228 | | | 2353 | 2354 | | 2091 | 1919 | 1611 | 2352 | 20023 |
| Spring barley | 1153 | 1152 | | 1027 | | 1078 | | | | | | 1021 | 1111 | 6542 |
| Spring oat | 1163 | 1165 | | 1020 | | 1114 | | | | 1021 | | 1070 | 1167 | 7720 |
| Sugar beet | 654 | 667 | | | 650 | | | | | | | | 655 | 2626 |
| Winter barley | 2079 | 2056 | | 1963 | | 2247 | | | | | | 2144 | 2331 | 12820 |
| Winter rapeseed | 1501 | 1489 | 1176 | 1274 | | 1541 | | 1802 | | | | 1369 | 1730 | 11882 |
| Winter rye | 1155 | 1140 | | 1176 | | 1300 | | 1301 | 1253 | | | 1164 | 1303 | 9792 |
| Winter wheat | 2276 | 2177 | | 2047 | | 2343 | | | | 2049 | | 2202 | 2466 | 15560 |

## 2.2. Crop field boundaries

Various radiuses have been proposed to identify a representative observation radius from 1 km to 20 km around the DWD phenological observation stations (Kowalski et al. 2020; Lobert et al. 2023; Tian et al. 2021). Following the DWD suggestions, we fixed this as an observation square box to a 5 km buffer around each station (Kaspar et al. 2015; Lobert et al. 2023). This buffer accounts for the limited geolocation accuracy of DWD stations, which are sometimes rounded to one decimal place, resulting in a maximum spatial uncertainty of 0.05°—approximately 5 km. Therefore, the 5 km buffer is consistent with previous recommendations to ensure the extraction of representative DWD observation data.

A Crop Type Map (CTM) of Germany (Blickensdörfer et al. 2022) was used to identify the available crop types located within a 5-km × 5-km square box around each DWD station (Figs.

1B-D). For each therein selected crop field, we used a two-step approach, following the suggestion by Lobert et al. (2023), to; first, remove the effect of field boundaries (e.g. hedge rows, or field borders, headlands, etc.) by discounting a 70 m inside buffer of the field, and secondly, removing an outside buffer of 40 m. Subsequently, each remaining effective field larger than 2 hectares area was selected as a candidate field .

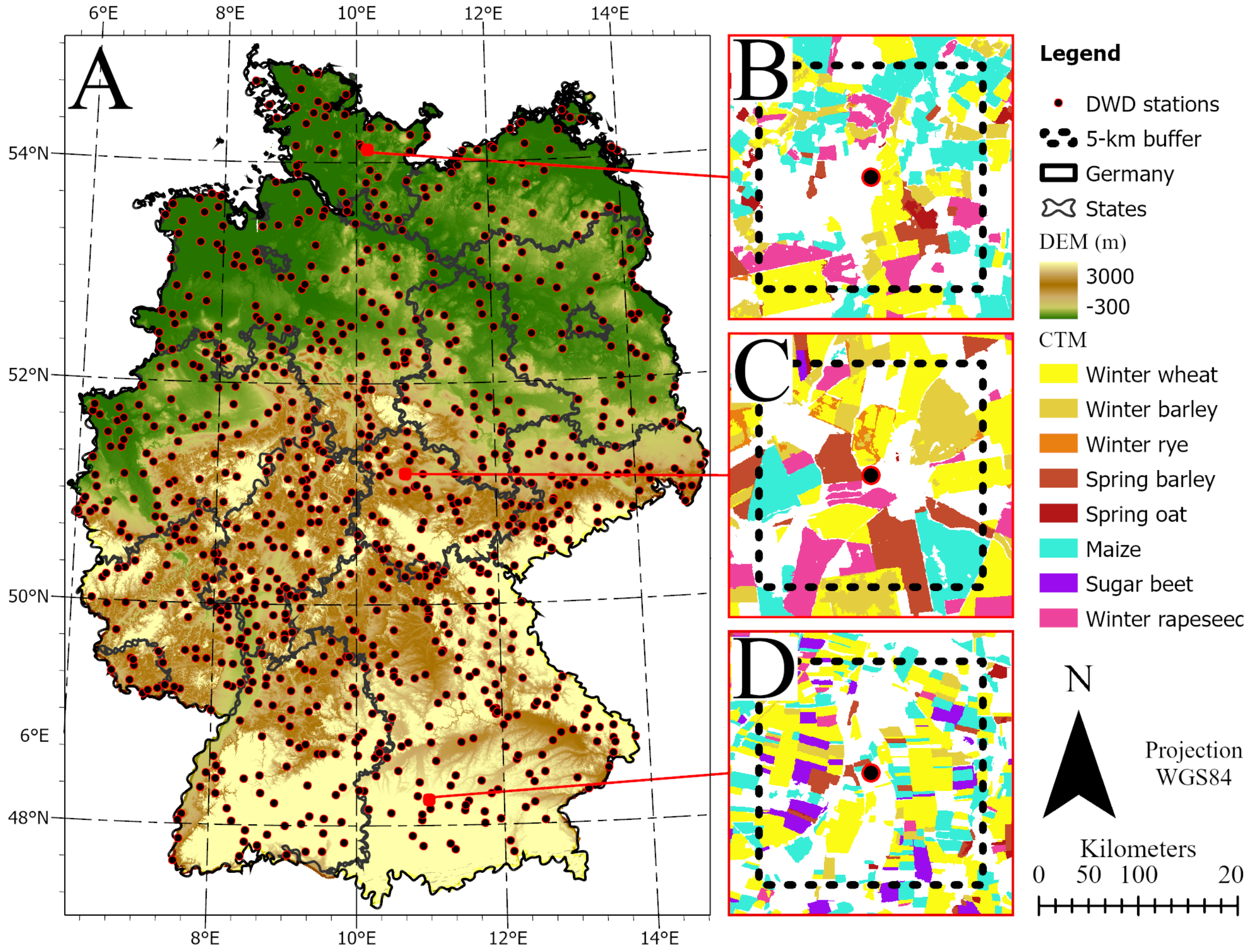

**Figure 1** Distribution of selected DWD stations on a digital elevation model (DEM; A) with a 5-km square buffers around three example stations (B-D; white patches indicate non-target areas) showing the Crop Type Map (CTM).

## 2.3. Remote sensing data and indices

### 2.3.1. Sentinel-1

The European Union's Copernicus Sentinel-1 (S1) mission is a constellation of two polar-orbiting satellites that carry a C-band synthetic aperture radar (SAR) instrument. The primary advantage of this satellite is the acquisition of imagery regardless of the weather conditions, which makes it ideal for land monitoring. We used the Sentinel-1 GRD FLOAT of both Sentinel-1A and Sentinel-1B with instrument mode of interferometric wide swath (IW) in Google Earth Engine (GEE). Mullissa et al. (2021) introduced a methodology to prepare analysis ready data (ARD) of SAR backscatter in GEE. Following their routine (c.f. https://github.com/adugnag/gee_s1_ard), we corrected SAR backscatter coefficient in the following order: i) additional border noise correction, ii) calibration to Gamma0, iii) speckle filtering (Multi-Boxcar), iv) orthorectification (terrain correction), v) terrain flattening (using NASA SRTM Digital Elevation based on the volume method), and vi) conversion to dB.

S1 acquires data with dual polarization of vertical transmit and vertical receive (VV; dB) and vertical transmit and horizontal receive (VH; dB) known as backscatter coefficients. Various studies suggested that VV backscatter coefficient is valuable for characterizing crop growth and development (Hu et al. 2024; Yang et al. 2024a; Yeasin et al. 2022) while VH backscatter coefficient is informative about crop biomass and moisture content (Löw et al. 2024) and less sensitive to factors like topography compared to VV (Yang et al. 2021). Additionally, some studies suggested indices based on VH and VV such as backscatter cross-ratio (CR=VH-VV) and backscatter ratio (PR=VH/VV) (Lobert et al. 2023; Schlund and Erasmi 2020). The Radar Vegetation Index (RVI=4/[1+VV/VH]) is a well-known index for monitoring vegetation dynamics (Mandal et al. 2020), and shows a potential to monitor crop growth (Haldar et al. 2022) and has a meaningful correlation with Leaf Area Index (LAI) (Pipia et al. 2019).

### 2.3.2. Sentinel-2

The European Union's Copernicus Sentinel-2 (S2) is a mission that consists of two polar-orbiting satellites equipped with multispectral instruments (MSI). S2 is well known for its ability to capture high-resolution imagery and makes it ideal for agricultural applications. We used harmonized Sentinel-2 MSI of level-2A (bottom of atmosphere) of both Sentinel-2A and Sentinel2-B in GEE to monitor crop phenology.

S2 acquires data from six optical bands of blue (10 m), green (10 m), red (10 m), near infrared (NIR; 10 m), with two short waves infrared (SWIR1 and SWIR2; 20 m), and four red-edge bands (re1, re2, re3, and re4; 20 m). The blue band is useful to discriminate between soil and vegetation (Tucker 1978). The green band is useful to asses plant health and vigor (Revill et al. 2019). The red and NIR bands are well known to provide valuable insights into crop health and growth stages (Revill et al. 2019), and SWIR1 and SWIR2 are effective to measure moisture content in both soil and vegetation (Liu et al. 2021b). The red-edge bands are known to help in vegetation classification; however, some studies suggested these bands provide valuable information for monitoring growth stages (Delegido et al. 2011; Kang et al. 2021). Although each band contains effective information about vegetation changes, calculated vegetation indices are proposed to enhance the sensitivity to vegetation changes, reduction of atmospheric effects, and improve signal to noise ratio (Huete et al. 2002). Therefore, we calculated and used various vegetation indices for different purposes as summarized in Table 2.

**Table 2** Multispectral indices used for crop phenology estimation in different studies.

| Class | Index | Abbreviation | Reference |
|---|---|---|---|
| Vegetation Health and Density Indices | Normalized Vegetation Index | NDVI | (Feng et al. 2024; Zamani-Noor and Feistkorn 2022) |
| | Enhanced Vegetation Index 2 | EVI2 | (Lobert et al. 2023) |
| | Green Normalized Difference Vegetation Index | GNDVI | (Feng et al. 2024) |
| | Green Chlorophyll Vegetation Index | GCVI | (Shrestha et al. 2023) |
| | Soil Adjusted Vegetation Index | SAVI | (Sitokonstantinou et al. 2023) |
| Water and Stress Indices | Normalized Difference Water Index | NDWI | (Senaras et al. 2024; Sitokonstantinou et al. 2023) |
| | Plant Senescence Reflectance Index | PSRI | (Senaras et al. 2024; Sitokonstantinou et al. 2023) |
| | Modified Chlorophyll Absorption Ratio Index | MCARI | (Senaras et al. 2024) |
| | Normalized Difference Infrared Index | NDYI | (Delegido et al. 2011; Zamani-Noor and Feistkorn 2022) |
| Atmospheric Corrected Indices | Atmospherically Resistant Vegetation Index | ARVI | (Vina et al. 2004) |
| | Wide Dynamic Range Vegetation Index | WDRVI | (Yang et al. 2022) |
| | Visible Atmospherically Resistant Index | VARI | (Shrestha et al. 2023) |

## 2.4. Auxiliary data

### 2.4.1. Climate data

We used daily maximum and minimum temperature, and precipitation from 2144 Meteostat stations across Germany (Fig. A1). We used the 'Meteostat' package in Python (https://github.com/meteostat/meteostat-python) to download the data covering the time period 2017 to 2021. Subsequently, we calculated Growing Degree Days (GDD) and cumulative Growing Degree Days (GDD sum) for each crop using average daily temperature (McMaster and Wilhelm 1997). The GDD and GDD sum are calculated from the beginning of the season, starting in the Fall (265th day of year) for both spring and winter crops following (Fu et al. 2014). We used base temperature of 4.5℃ for winter crops of wheat, rapeseed, rye, and both spring and winter barley, 10℃ for maize, 0℃ for spring oat (Center 2001), and 1℃ for sugar beet (Holen and Dexter 1996). We also calculated Diurnal Temperature Range (DTR) to measure day and night temperature difference for a better understanding of the condition of the plant phase during the day (Huang et al. 2020). Lastly, we used cumulative precipitation (precipitation sum) as a measure for the water availability of crops (Le Roux et al. 2024).

#### 2.4.2 Elevation data

After the Copernicus Digital Elevation Model (CDEM) was released in 2019 which is the currently most recent DEM dataset in 30 m resolution, various studies have investigated its accuracy (Guth and Geoffroy 2021; Li et al. 2022a; Liu et al. 2023). The wide consensus is that this dataset is considered very reliable. However, like any DEM dataset, it is also affected by trees and buildings. In solution, Hawker et al. (2022) used a Machine Learning (ML) model to remove buildings and tree height biases from CDEM named Forest And Buildings removed Copernicus DEM (FABDEM) to overcome this limitation. Some recent studies have evaluated the accuracy of FABDEM considering it highly reliable for bare land terrains (Dandabathula et al. 2023; Marsh et al. 2023). Since hedgerows and agroforestry are common features of the agricultural landscape in Germany, the FADEM dataset is used (Fig. A2).

Altitude (i.e., elevation) is the major factor that changes the climate factors and consequently the plant growth, while slope and aspect affects the diversity and density of plants (Marini et al. 2007; Singh 2018). Therefore, we used altitude as well as slope and aspect in this study. We used FABDEM dataset in GEE and calculated slope using "ee.Terrain.slope" and aspect using "ee.Terrain.aspect" commands in GEE.

### 2.5. Data preprocessing

Temporal revisiting of S1 is 2-4 days and of S2 is 3-5 days. This frequency makes it ideal for vegetation monitoring. We upscaled the S1 spacing of 20 m so that it matched the resolution of the S2 bands (all S2 bands also upscaled to 20 m to match SWIR1 and SWIR2 resolution). Due to frequent cloud cover in Germany, S2 images had to be masked (images with cloud probability <75% followed by masking cloud, cirrus, and cloud shadows). We used the S2 cloud probability dataset in GEE to mask out clouds. To remove undesired noise and artifacts, the "lowess" smoother was used for time series smoothing using the Python Package "tsmoothie" (see https://github.com/cerlymarco/tsmoothie) and selected a smooth fraction of 0.03 based on visual interpretations, as suggested by Lobert et al. (2023). When applying smoothing with a high fraction for LOESS, a selection of 0.03 proved effective in preserving the peak values of optical, SAR, and climate parameters during growing seasons. The processed variables were then resampled to daily resolution using linear interpolation to match the time stamps of DWD phenology reference data. Each phenology phase is associated with the relevant satellite and climate data for each DWD station. If a phenological stage was not available, the corresponding date is labeled as background (-1). The DWD stations are not geotagged (i.e., not linked to a unique geolocation or parcel). Therefore, we assumed that the median values of S1 and S2 features across the selected fields within the bounding 5x5 $km^2$ boxes around the DWD stations is representative. For the climate data, we calculated the distances between phenology stations and climate stations, and we selected the ten nearest climate stations to use inverse distance weighting (IDW) with the aim to obtain the necessary climate variables at each DWD station in question. As an example, Fig. 2 shows the preprocessed data for winter wheat at a sample DWD station in Hohenbachen in Bavaria for the 2020-2021 growing season.

**2.6. Machine learning model**

Tree-based gradient boosting (GB) learning algorithms recently got attention in many fields of science and show promising performances in remote sensing of vegetation and land use land cover classification (Gao et al. 2023; Zhang et al. 2021). Due to the structure of GB methods, they may deal with highly nonlinear interrelations between predictors and response variables in the form of an ensemble of weak predictions (Chen and Guestrin 2016). Various GB models have been introduced and among them, we selected the LightGBM classification ensemble tree model which has a multitude of features such as scalability, efficiency, and handling large-scale data (Ke et al. 2017). The LightGBM was executed in Python using the LightGBM package (see https://github.com/microsoft/LightGBM).

Although GB models are efficient, they are problematic in finding the global optimum. Therefore, we employed the hyper-tuning algorithm of Optuna in Python to find the best parameters of the LightGBM classification model (Akiba et al. 2019). The Optuna is a hyperparameter optimization framework that uses the advanced sampling technique of a Tree-structured Parzen Estimator (TPE) sampler. The TPE sampler can make a multivariate suggestion for parameters similar to advanced Bayesian Optimization and HyperBand (BOHB) algorithm (Falkner et al. 2018).

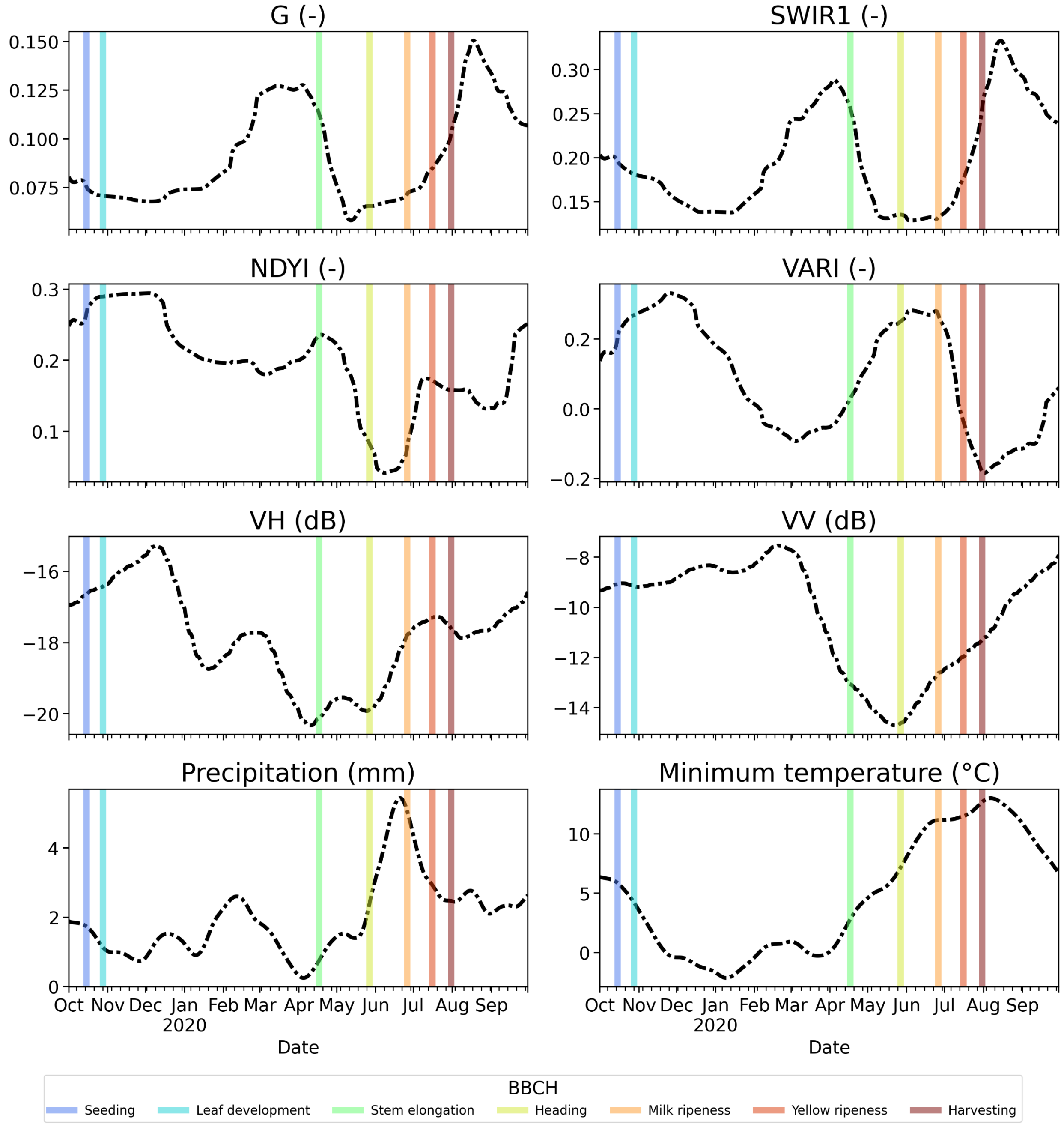

**Figure 2** Time series of RS and climate parameters for winter wheat at different BBCH stages in 2020 growing season at Hohenbachen in Bavaria; G: Green band, SWIR1: Short Wave InfraRed 1, NDYI: Normalized Difference Infrared Index, VARI: Visible Atmospherically Resistant Index, VH: Vertical transmit and Horizontal receiver, VV: Vertical transmit and Vertical receiver (VH; dB)

## 2.7. Training and evaluation

The labeled dataset displayed an imbalance within each crop (Table 1). Thus, we trained independent models for each crop, individually. Further, there were significant imbalances within each BBCH growth stage in each crop. For instance, in ideal conditions, there were at least 358 background labels for winter wheat that observed 7 growth stages. Two different strategies were used to adjust the influence of imbalances. The first strategy involved class weights, directing the model to pay special attention to rarer events (BBCH phases). The second strategy is nested cross-validation. We made a random k-fold split into 10 random folds following previous studies (Lobert et al. 2023; Ziegler et al. 2020); for each split, 9 folds for training and 1 for testing were used for the outer loop to evaluate model performance. The training set is further divided into 10 folds in the inner loop, with 9 used for training and 1 for validation. The average accuracy for the inner folds was the corresponding loss function for hyperparameter tuning in each test fold. In total, we trained and fine-tuned 100 models that would yield 10 different models corresponding to each test fold from the inner folds. Then, an average ensemble approach was used for the final prediction in the test phase. It helped to reduce bias in performance estimates, provided a more representative sample in each fold, and addressed the sizable imbalances between the phenological stages. Spatial cross-validation effectively mitigates spatial autocorrelation, but it may produce more conservative performance estimates and potentially underestimate model performance for within-domain applications (Wadoux et al. 2021). Since our method is intended for use within the environmental domain of Germany, random cross validation remains appropriate (Kattenborn et al. 2022; Lobert et al. 2023).

For the hyper-tuning of LightGBM parameters, *n_estimators*, *num_leaves*, *min_data_in_bin*, *min_child_samples*, and *early_stopping_round* were used while the objective was multiclass classification and an objective function (Section 2.8). To ensure we find the best parameters, for each inner fold, we did 50 trials for hyper tuning. This resulted in 10 different models for each test fold. Early stopping on the validation dataset helps prevent model overfitting and aids in finding the global optimum.

The proposed framework for the estimation of phenology relies on the CTM, geospatial location (latitude and longitude), DEM and its derivatives, remote sensing parameters, and data climate information (Fig. 2). These parameters were used to train, validate, and evaluate a classification

ML model to extract the onset of each phenology stage in the day of the year. To help the model find the best feature set, we used the Optuna for feature selection as well. For this, we used the TPE sampler, a Bayesian optimization algorithm, which has been ensured to produce reasonable results for feature selection processes (Yang et al. 2024b). To make proposed feature selection among S1, S2, climate, DEM, and geospatial location features more practical, we collected each feature in different groups and first optimized for the groups (Table 3) in total 44 features. If the group was selected, then in each group we searched for each feature that has more impact on loss function. If the feature and feature group help the model to decrease the loss function, the feature and feature group is selected for test phase. Then we standardized the features selected between crops (Algorithm 1). We found that 50 trials would be enough to find the best feature group and feature set.

Table 3 Full combinatory feature groups for selecting input features.

| **ID** | **Group** | **Feature set** | **# Feature** |
|---|---|---|---|
| 1 | SAR | VH, VV | 2 |
| 2 | Optical | B, G, NIR, R, SWIR1, SWIR2 | 6 |
| 3 | Red edge | re1, re2, re3, re4 | 4 |
| 4 | Climate | Maximum Temperature (tmin), Minimum Temperature (tmax), Precipitation (prcp) | 3 |
| 5 | Vegetation Health and Density Indices | NDVI, EVI2, GNDVI, GCVI, SAVI | 5 |
| 6 | Water and Stress Indicators | NDWI, PSRI, MCARI, NDYI | 4 |
| 7 | Soil and Atmospheric Correction Indices | ARVI, WDRVI, VARI | 3 |
| 8 | SAR Based Vegetation Indices | RVI, PR, CR | 3 |
| 9 | Climate indices | GDD, GDD sum, DTR, Precipitation sum (prcp sum) | 4 |
| 10 | Time features | Season, Month, Day of week, Day of month, Day of year* | 4 |
| 11 | Geospatial features | Latitude, Longitude | 2 |
| 12 | Elevation data | Altitude, Slope, Aspect | 3 |
| Total | | | 43 |

*To calculate the date of phenology stages, day of year is needed; however, it is not an input data for the prediction model.

**Algorithm 1.** Feature optimization and selection

**Input**:

- Dataset $D$ with features from geospatial, climate, and remote sensing sources for crops.
- Number of folds $K$ = 10.
- Loss function $L$ (Section 2.8).
- Target: Predict phenological stages.

**Output**:

- Optimized feature set $F_{opt}$ per crop.
- Standardized feature set $F_{std}$ across crops.

**Feature sampling and model training: do:**<br>
Split $D$ into $K$ = 10 folds for each crop $C$.<br>
**For each fold $K$ (1 to 10): do**:<br>
Train model $M_k$ using all features in $D$.<br>
Record feature set $F_k$ and loss $L_k$.<br>
**end do**<br>
**Feature importance calculation: do:**<br>
**For each crop C: do**:<br>
**For each feature $f \cup F_k$: do**:<br>
Compute frequency $N_f$ as number of occurrences across folds.<br>
Compute average loss $L_f$ for folds including $f$.<br>
Calculate importance $I_f = N_f / L_f$.<br>
**end do**<br>
Rank features by $I_f$.<br>
**end do**<br>
**Optimized feature selection: do:**<br>
**For each crop C: do**:<br>
Select top features from ranked $I_f$ maximizing accuracy.<br>
Assign selected features to $F_{opt}$ $(C)$.<br>
**end do**<br>
**Standardized feature set creation: do:**<br>
Define $F_{std}$ based on high $I_f$ among crops.<br>
Retrain models with $F_{std}$.<br>
Evaluate performance:<br>
Compute $R^2$ and $MAE$.<br>
**Sensitivity analysis (optional): do**:<br>
**For crop (e.g., winter wheat): do**:<br>
Remove location features from $F_{std}$.<br>
Retrain and evaluate:<br>
Compute $R^2$ and $MAE$.<br>
Remove elevation features from $F_{std}$.<br>
Retrain and evaluate:<br>
Compute $R^2$ and $MAE$.<br>
**end do**

## 2.8. Performance metrics

The ML model performance to predict phenology stages is evaluated by two metrics of the coefficient of determination ($R^2$) and Mean Absolute Error (MAE) as follows:

$$R^2 = 1 - \frac{\sum_{i=1}^{n}[y_i - \hat{y}_i]^2}{\sum_{i=1}^{n}[y_i - \bar{y}]^2} \quad ; \quad 0 \leq R^2 \leq 1 \qquad (1)$$

$$MAE = \frac{\sum_{i=1}^{n}|y_i - \hat{y}_i|}{n} \quad ; \quad -\infty < MAE < \infty \qquad (2)$$

Where, $y_i$ and $\hat{y}_i$ are observed and estimated values of phenology date [day of year], respectively, $\bar{y}$ [day of year] is the average value of observation, and $n$ denotes the number of observations.

We proposed an objective function based on these metrics in the form of $(1 - R^2) \times MAE$ to be minimized. This objective function provides a helpful gradient (direction) for Optuna and was shown to shape the objective function space favorably for the searching of a global optimum, while ensuring matching units (i.e., day of year).

## 2.9. Processing environment

In this study, most of the computations, analysis, preprocessing and postprocessing of the proposed framework have been established in the GEE environment. Then the postprocessed data was downloaded for each crop and the LightGBM model was trained, and the analysis of the model results was done using Python.

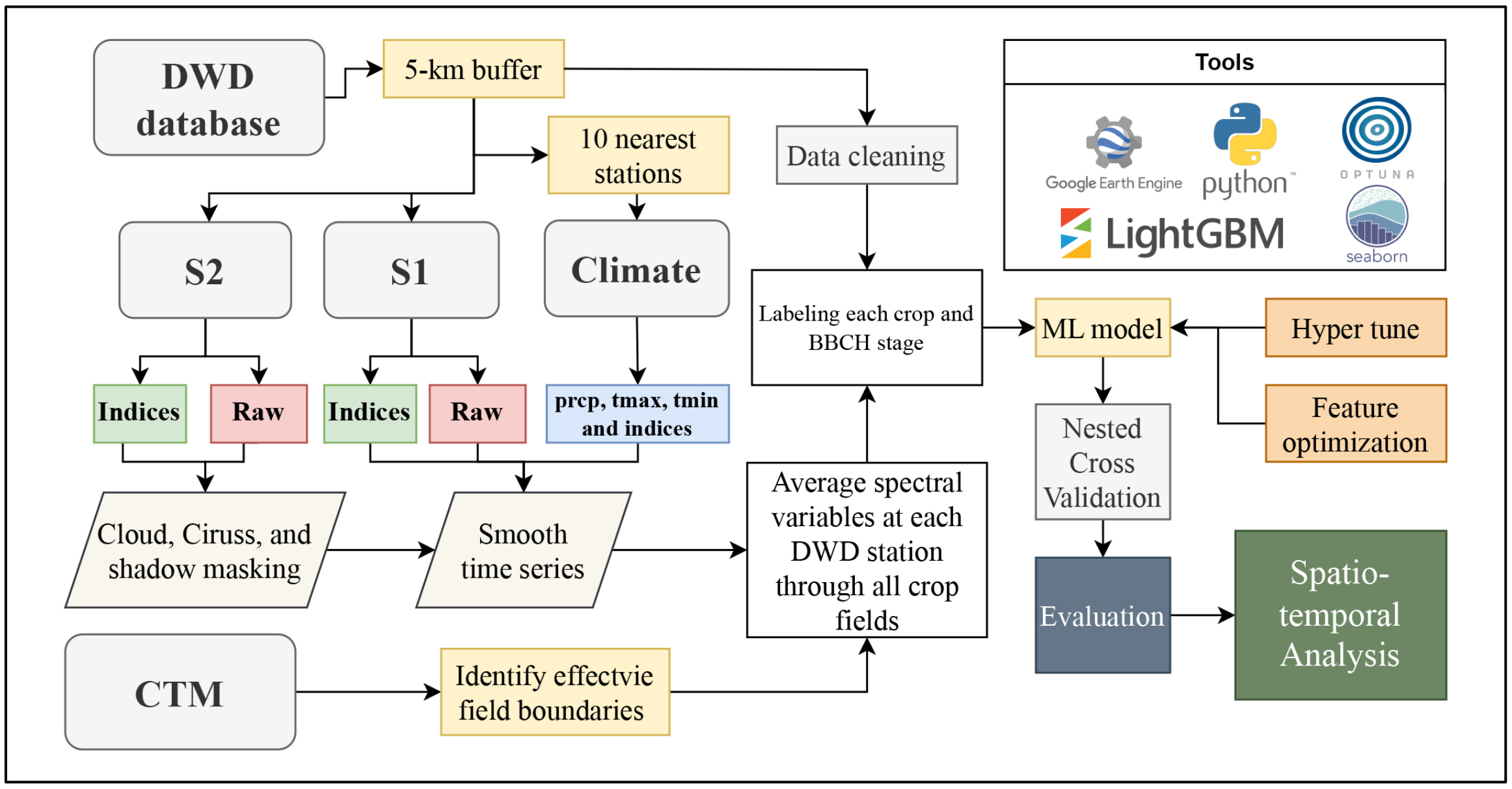

**Figure 3** Flowchart of the proposed framework. All abbreviations are defined in the text.

# 3. Results

This study proposed a framework to detect phenological stages of eight crops for 13 different stages across Germany between 2017 and 2021. By fusing optical, radar, and climate datasets, this study evaluated the accuracy of the ensemble models for each crop and phenology stage (Section 3.1) and assessed the impact of each feature set and individual feature on crop phenology modeling using the ML model (Section 3.2). Then the spatio-temporal variability and transferability of the model is evaluated as well (Section 3.3).

## 3.1. Feature selection

Selecting features in this study was based on the model accuracy with the aim to find the best combinations of features and standardizing them between crops from different data sources (Table 2, Algorithm 1). We trained an individual model for each crop. Each was trained for 10 different unseen folds resulting in 10 different distinct models with different feature sets. To find the best feature set for each crop, we selected features that occurred most frequently in the folds and the loss function value (which is proposed for hyper tuning, Section 2.8). Therefore, we calculated the importance for each selected feature based on the number of repetitions in the folds divided by the loss that this feature set finally giving for prediction. Then, we selected features that have the highest feature importance and led to higher accuracy of detecting and estimating the onset of each phenological stage (Fig. 4). For each crop, selected features are different, while in all crops, some features are essential and repeated frequently. Thus, the most repeated features are selected as an standardized features for all crops.

Among the most important features, the GDD sum (averaged importance between crops ~ 7.7) stands out as the top contributor for identifying crop growth stages followed closely by DTR (7.5). Geospatial features also play a pivotal role, with longitude and latitude showing high feature importance (7.2 and 6.9, respectively). Altitude (6.4) and its derivative aspect (6.3) follow, providing additional insights, while precipitation sum (6.1) further enhances the model's predictive power. Slope (5.5) and GDD (5.5) also demonstrate notable importance, reinforcing the frequent selection of geospatial features and elevation-related groups by the feature optimization algorithm. Even though altitude implicitly informs the model about spatial temperature differences, the distinct significance of climate indices like GDD sum and DTR remain evident. From RS data, parameters vary between crops, with the most consistently repeated being VV (4.3), PR (3.2), and RVI (3.1) indices from the S1 satellite, while no parameters from optical satellites are selected. The analysis also showed that without climate data or Sentinel-1 inputs, the model is not capable of accurately predicting phenology, underscoring their indispensable roles. Other RS parameters are less frequently repeated across different crops.

To standardize the feature set and enhance its applicability for future research, we retrained our model using the most frequently occurring features of latitude, longitude, altitude, slope, aspect, DTR, GDD sum, GDD, precipitation sum, VV, PR, and RVI. The results showed a slight decline in modelling performance, with a 2.3% decrease in $R^2$ (Fig. A3) and a marginal rise in MAE of only 0.5% (Fig. A4). These findings suggest that the differences are not significant and the proposed method for feature importance analysis was successful to find the most important features. Additionally, we examined the impact of static features on model accuracy and error, focusing specifically on winter wheat (Figs. A5 and A6). Our findings suggest that location and elevation influence the early stages of growth and help the model capture geospatial differences caused by varying planting dates across regions across Germany. However, these features are not essential for the proposed model to identify phenological stages. Consequently, model accuracy decreases slightly when these features are removed ($R^2$: location feature set by 9.3% and elevation feature set by 2.3%), while error metrics showed a slight increase in the mean absolute error (MAE: location feature set by 1.8% and elevation feature set by 1.3%, respectively).

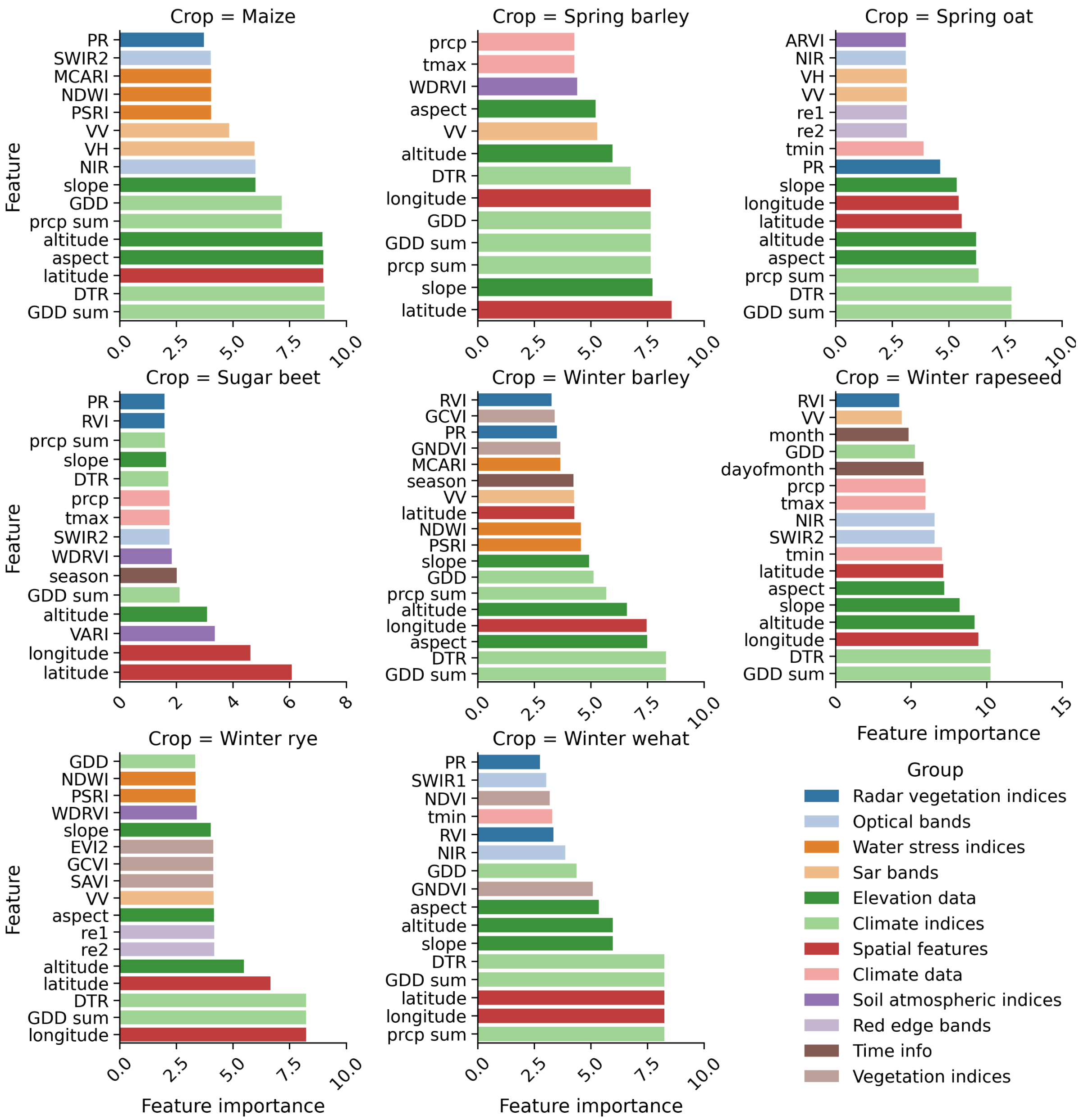

**Figure 4** Selected features for each crop based on feature optimization.

## 3.2.Model efficiency

The phenology estimates for each crop are shown in Fig. 5. The proposed method is quite consistent across different crops with an average MAE of 5.8 days, although there is a variation between different phenological stages. Model error is closely tied to the number of observations; specifically, a higher number of observations results in average MAE values below 6 days for maize, spring barley, spring oat, winter barley, winter rapeseed, winter rye, and winter wheat. In sugar beet, which has lower observation numbers, exhibit average MAE values exceeding 6 days. The goodness of fit, expressed as $R^2$ for all phenological stages together, is above 0.97. Hence, the proposed method performs well during the entire growing season, though the linear regression red lines show a slight deficiency in the predictions of individual BBCH stages.

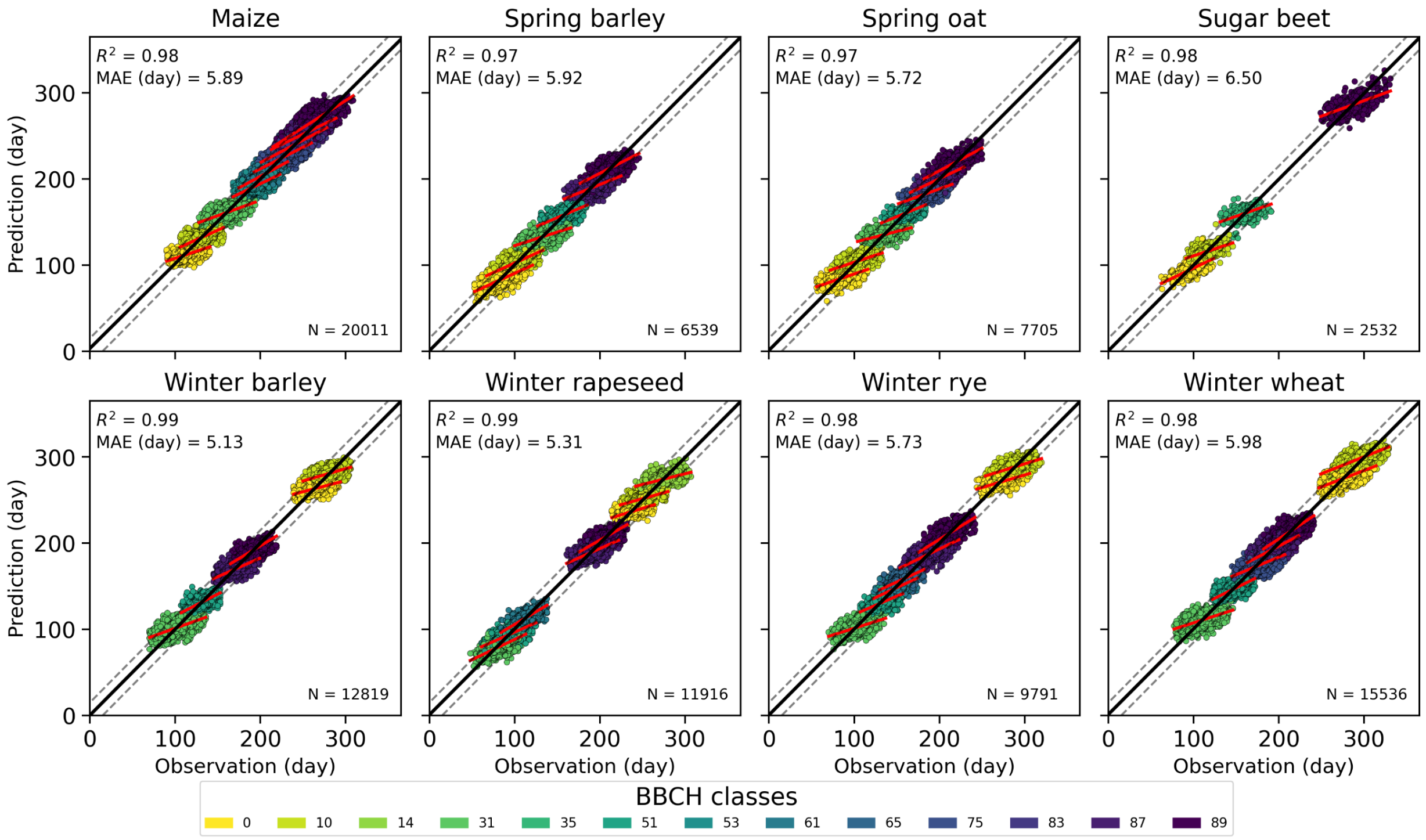

**Figure 5** Scatter plot of predictions of unseen data for each crop and BBCH. The dashed black lines are the deviation of ±15 days from the 1:1 line. The red lines are the linear regressions between observations and prediction for each BBCH stage.

To address this, we evaluated the effectiveness of the proposed model in predicting and detecting the phenology of each crop, assessing the model's error and accuracy across different crops and BBCH stages (Fig. 6). Model performance showed some variation across crops and stages by MAE, $R^2$, and percentage of the predictions falling within ±6 days of the observed value, while in general, the model accuracy, $R^2$, is increasing and, the model error, MAE, is decreasing by crop development, except for spring barley and sugar beet.

At early stages of seeding (BBCH 0) and emergence (BBCH 10), the model usually had a MAE of lower than 6 days with more than 63% of predictions lie within ±6 days, except for winter rye and winter wheat. Shooting (BBCH 31), as the third early stage, contributed to the highest MAE in all crops, with a value higher than 6 days for all crops. In winter barley, the MAE of shooting was 7.5 days, while only 56% of the predictions were within ±6 days from observations. Similarly, for maize, the MAE for BBCH 31 was 6.8 days, with only 55% of the predictions lying within the ±6-days range. Therefore, the early stages (BBCH 0-31), are problematic for the model to detect the exact date of phenological stages except for maize and rapeseed.

The middle stages of crop growth (BBCH 51–65) show the best results and were similarly good for all crops. For instance, winter rye had the lowest MAE<4 days for beginning of flowering (BBCH 61) with 76% of predictions within ±6 days of the observations. The heading stage (BBCH 51) showed the lowest MAE<4 days in winter barely with 79% of predictions within ±6 days of the observations. In maize, the tip of tassel visible (BBCH 53) resulted in a MAE of 5.48 days. General flowering (BBCH 65) for winter rye with MAE of 5.74 days show that the model could find a good relationship with input data and this phenological stages.

For the later stages of crop phenology (BBCH 75–89), MAE remained below 6 days for most crops, except for maize and sugar beet, which showed higher MAE (up to 8–10 days). Excluding these, MAE decreased and $R^2$ increased toward the end of the season, particularly in winter crops like winter barley, rapeseed, rye, and wheat, where BBCH 89 had a low MAE (<4.5 days) and a high $R^2$ (>0.55)—similar to the MAE at the heading stage (BBCH 51) but with an improved $R^2$. Spring barley and oat displayed a comparable pattern, though their $R^2$ was slightly lower at 0.5–0.6; while sugar beet lacked a distinct trend, and maize showed an upward trend in both MAE and $R^2$ as growing season progressed toward its conclusion.

As some DWD stations report geolocations with only one decimal degree of precision, the 5 km buffer (Section 2.2) was chosen to account for this limitation. To further explore the implications of this positional accuracy on our modeling framework, we conducted a specific analysis, the results of which are visualized in Figs. A7 and A8. Our evaluation revealed that only 25% of stations provide latitude or longitude with one decimal place. By excluding stations with one-decimal precision in either latitude, longitude, or both, and modeling of the station with high geolocation precision, we observed that the core modeling approach remained robust and largely unaffected. Specifically, model performance showed modest improvements: $R^2$ increased by 0.01, and MAE decreased by an average of 0.12 (days). These results suggest that the impact of geolocation precision on model uncertainty is negligible, reinforcing the reliability of our approach across varying levels of coordinate accuracy.

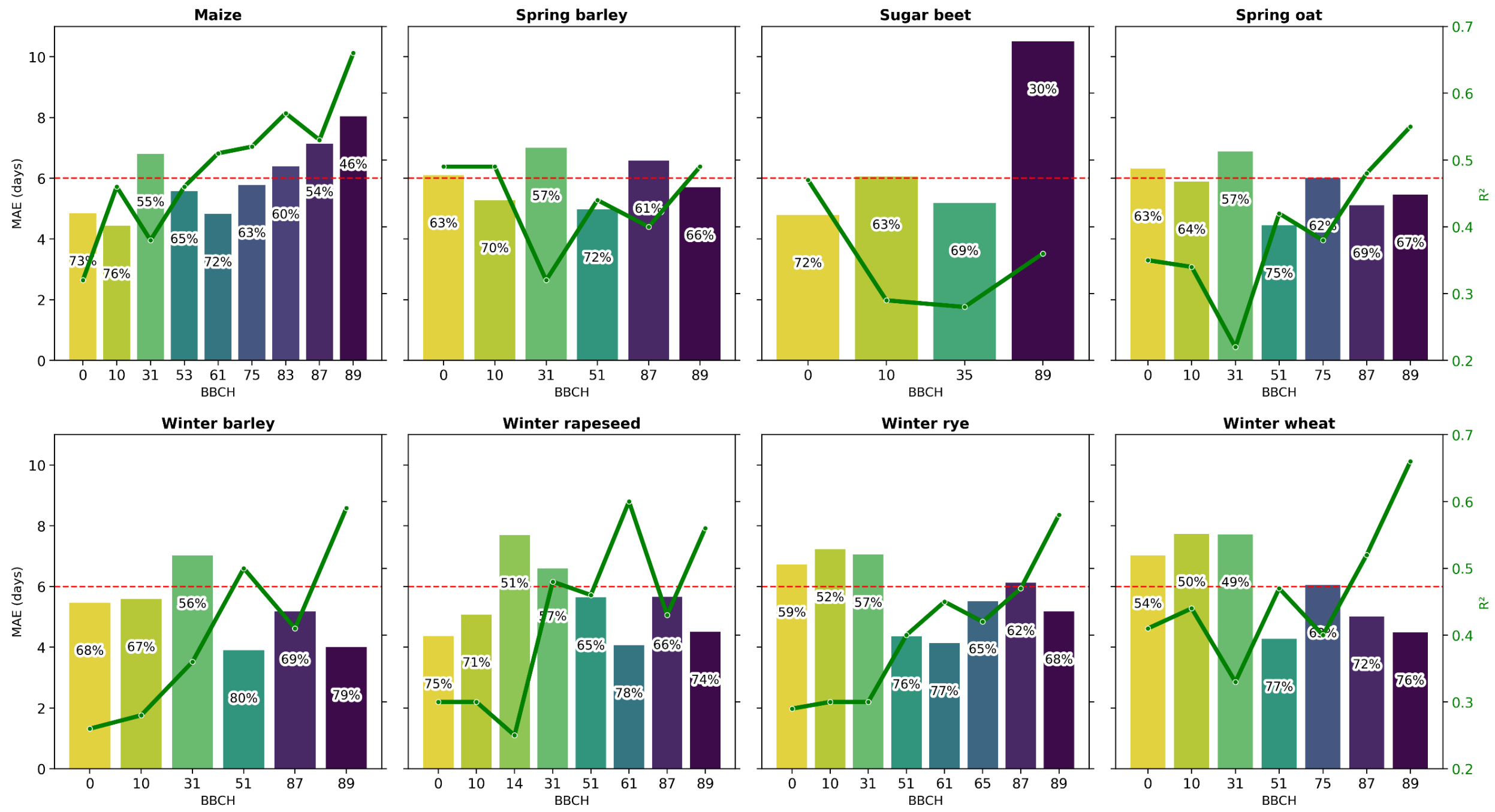

**Figure 6** Bar plot of the MAE (on left vertical axis) and line plot of $R^2$ (on right vertical axis) for each crop in each BBCH. The dashed red line is a deviation of ±6 days from the prefect prediction. The percentage shows the proportion of the prediction that has differences with observation within ±6 days.

## 3.3.Spatio-temporal evaluation

We studied the spatial patterns of the differences between prediction and observations, where we define this as residual = prediction – observation for each BBCH. In Figures 7 and 8, this is shown for the start and the end of the season, respectively. These stages were chosen primarily due to numerical reasoning: the early stages exhibit the lowest accuracy, while the later stages demonstrate the highest accuracy. For various crops, different spatial patterns in model performance can be seen for seeding stage; however, in general the model underestimates the day of seeding for all crops. The immediate in estimations (i.e., underestimations), for ranges higher than 4 days and lower than -4 days, are different for each crop as a percentage, which is for maize (-16%), spring barley (-19%), spring oat (-14%), sugar beet (-5%), winter barley (-22%), winter rapeseed (-13%), winter rye (-16%), and winter wheat (-29%). Although the model underestimates the seeding stage, there is no spatial pattern to this. For the late stage of harvesting, the model exhibit overestimation and delay in estimations for maize (32%), spring barely (7%), spring oat (9%), winter barley (8%), winter rapeseed (13%), winter rye (5%), and winter wheat (14%) except for sugar beet (-4%). In general, there is no spatial pattern towards local clustering which confirms our nested cross validation helped model to generalize and preserve predictive power over space by randomly choosing the stations.

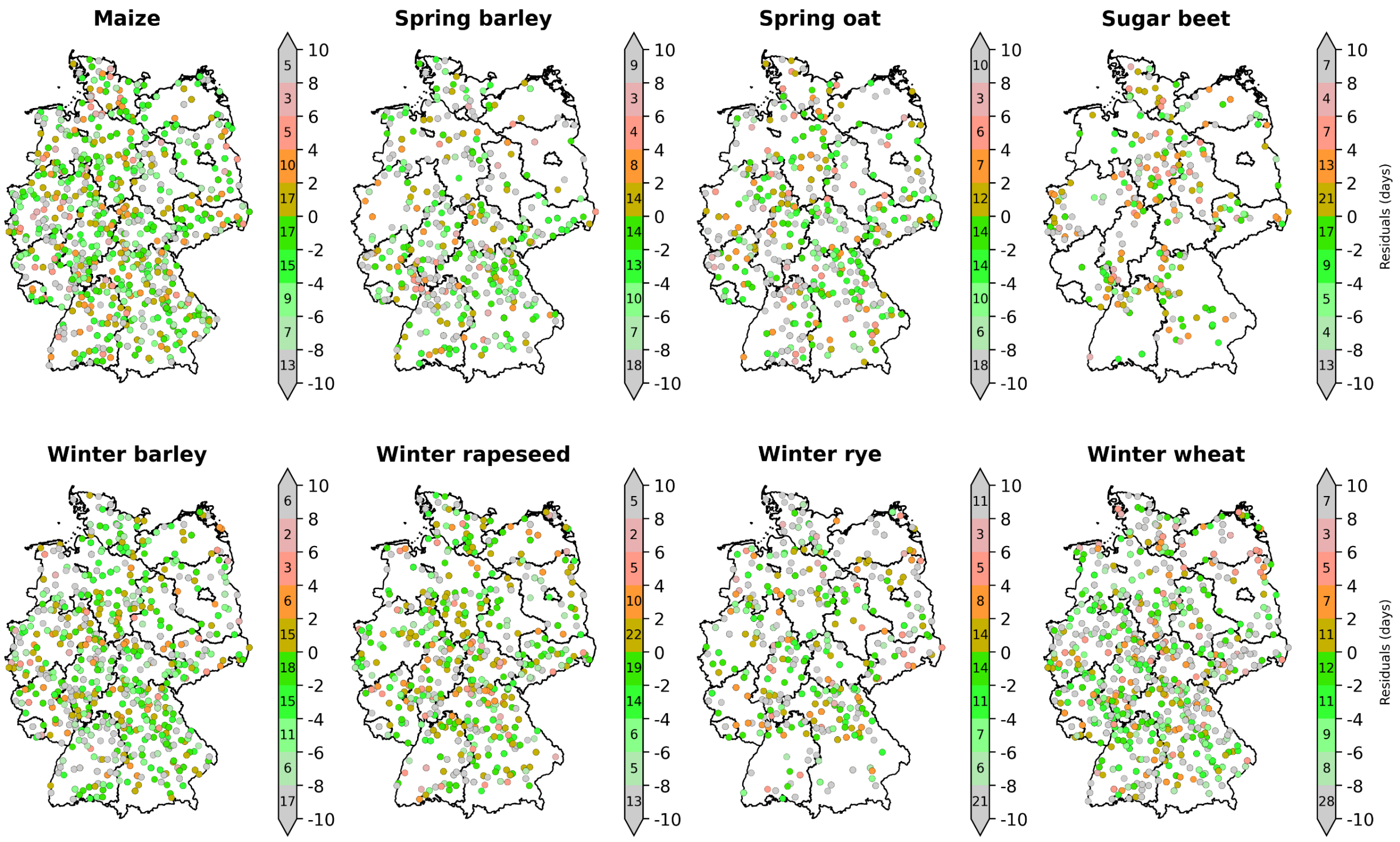

**Figure 7** Spatial difference between estimated and observed phenology for all crops at seeding stage across Germany. The color shows the number of days between estimated and observed values, and the numbers represent the concordant percentage of data in each range.

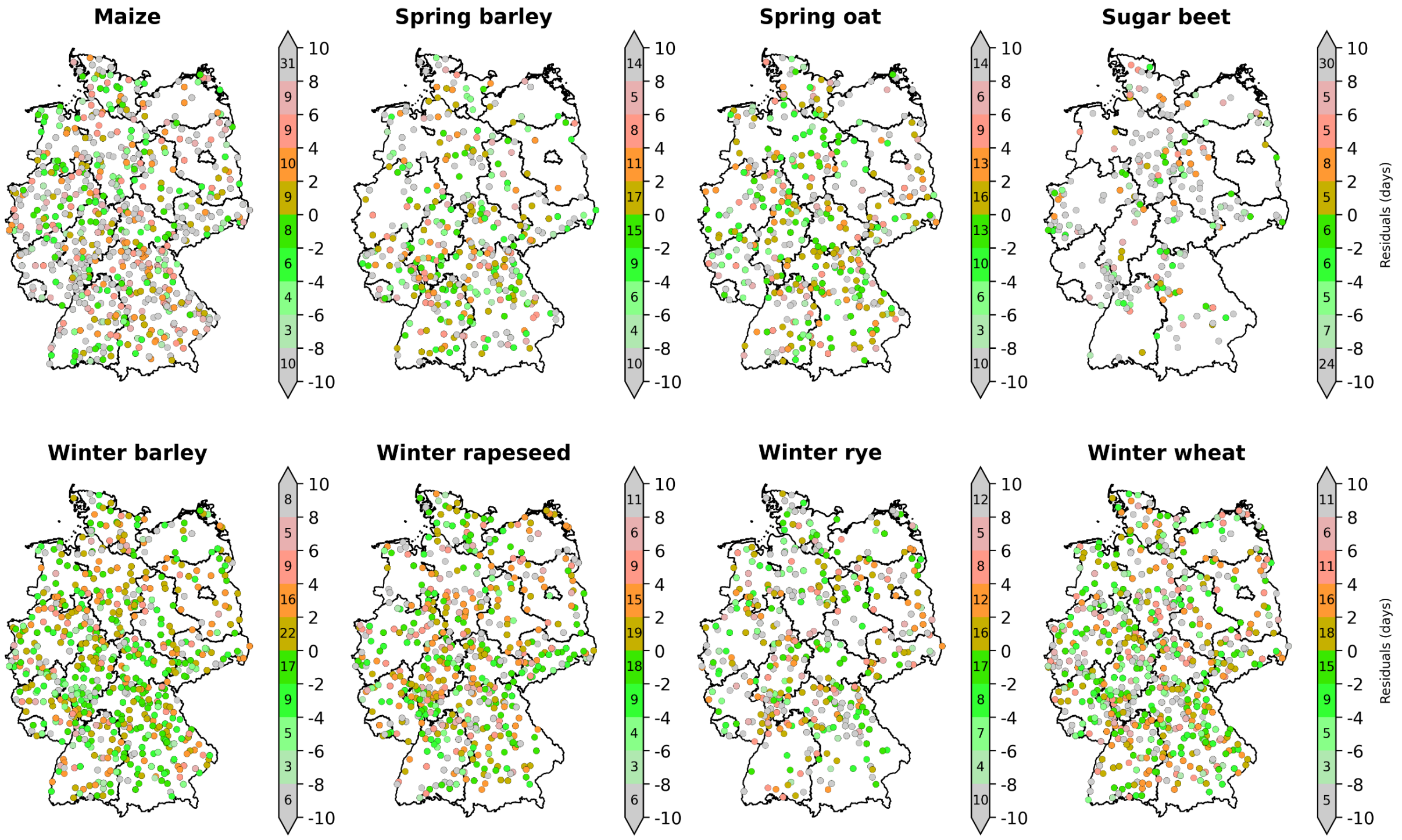

**Figure 8** Spatial difference between estimated and observed phenology for all crops at harvesting stage across Germany. The color shows the difference amount between estimated and observed values and numbers in each color represent the percentage of data in each range.

The temporal transferability of the models, as shown (Fig. 9), reveals distinct patterns between summer and winter crops. Summer crops like maize, spring barley, sugar beet, and spring oat have lower MAE averages in 2018 and 2020 (around 4–5 days) compared to the whole years average (6–7 days), due to drought in 2018 and higher temperatures in 2020, which shortened the growing season and reduced variability. However, in other years like 2017, 2019, and 2021, their MAE aligns with or exceeds the average, indicating weaker generalization across diverse weather patterns. Conversely, winter crops—winter barley, rapeseed, rye, and wheat—show consistent MAE (4–6 days) across all years, suggesting robust performance across temporal contexts due to their longer, more stable growing seasons. Overall, the temporal transferability analysis underscores the need for crop-specific model adjustments to account for the distinct responses of summer and winter crops to environmental conditions.

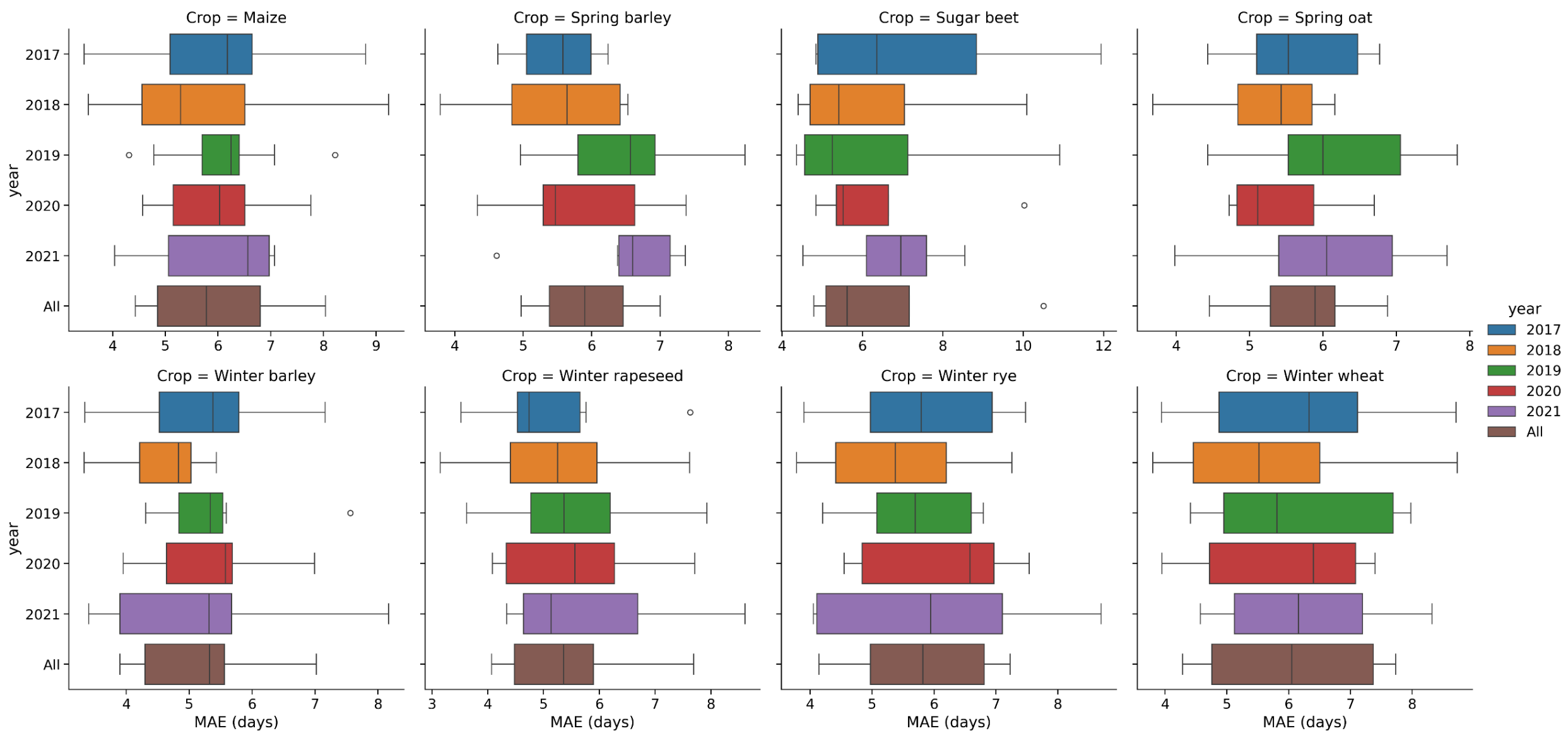

**Figure 9** Differences between MAE (days) of model predictions for each year and each crop in whole stages together. All in the year means for whole years together.

# 4. Discussion

## 4.1.Feature selection

We discovered that meteorological factors are crucial for precisely identifying phenological stages, aligning with the traditional view that climate elements, particularly temperature, play a dominant role in regulating phenological progression (Estrella et al. 2007; Hatfield and Prueger 2015; Patel and Franklin 2009). The GDD sum, DTR, and GDD are known parameters in phenology studies that control development rates, plant growth, photosynthetic efficiency, amongst others (Aslam et al. 2017; Hu et al. 2019; Jackson 1966). Precipitation sum as the source of water is a central factor for crop growth, where rainfed agricultural lands are predominant in Germany (Wu et al. 2023) which is well aligned with Hollinger and Angel 2009 study, since though precipitation effect is indirect, more affecting the plant growth through plant available soil water content.

Combining geospatial features and elevation data with climate and S1 data proved to be an effective set of features, while geospatial features demonstrate high explanatory power. This could be attributed to their ability to clarify the connections between geospatial variations in climate, photo period, management practices, soil fertility, and their impact on crop development, consistent with findings from previous studies (Eyshi Rezaei et al. 2017; Keenan et al. 2020; Zheng et al. 2016), and helped the model to better map phenological stages and account for localized management practices across Germany such as irrigation in the north (Baroni et al. 2019; Lobert et al. 2023). Singh (2018) and Mei et al. (2018) discuss slope, aspect, and altitude have critical role in regulating three key elements for crop development of soil temperature, soil moisture, and sunlight exposure.

The S1 features were also critical to accurately detect crop phenology which might stem from the cloud resilience and dense time series of S1 in comparison to S2 and aid in identifying critical moments when crops transition between phenological stages. This aligns with previous researches suggesting that, on average, SAR data tends to perform slightly better than optical data (Lobert et al. 2023; Meroni et al. 2021; Veloso et al. 2017), although other studies have proposed the opposite (d'Andrimont et al. 2020; Mercier et al. 2020). Additionally, previous studies confirmed backscatter ratio (PR) is a parameter that responds to crop morphology and has high correlation with NDVI (Hu et al. 2024), in which VH polarization is sensitive to the horizontal structure of crops (e.g., leaves) and VV is sensitive to vertical structures (e.g., stems) (Hu et al. 2024; Schlund and Erasmi 2020), which shows this study by selecting VV as a valuable factor is well-aligned with previous studies. RVI is also known as a biomass indicator from radar data that can help to identify the onset of greenness in spring and senescence in summer and fall (Haldar et al. 2021).

### 4.2.Model efficiency

It is a well-established fact that the number of observations significantly influences ML models (Beleites et al. 2013). This influence becomes particularly clear when comparing model accuracy across different scenarios. For instance, in the case of sugar beet, which had the smallest number of observations among the crops studied, the model's accuracy was noticeably lower compared to other crops that shared a similar set of features but had more data points. This suggests that insufficient data can limit a model's ability to learn and generalize effectively. However, the impact of data volume is not the only factor at play—each feature set, or group of input variables, may have a distinct influence on the model's performance depending on the specific phenological stage of the crop being analyzed. For example, model inaccuracies in start of the seasons may be related to the late time of seeding of these crops in fall and the absence of crop cover during the seeding stage, coupled with minimal cover crop presence during emergence, which may result in a significant proportion of soil signals being detected by radar sensors (Lobert et al. 2023; Veloso et al. 2017; Zeng et al. 2020). Lobert et al. (2023) study, which had combined radar backscatter coefficients and optical raw band data with CR and EVI in whole Germany for winter wheat, reported MAEs of higher than 7 days for the early stages (close to 10 days for BBCH 31); while Schlund and Erasmi (2020) suggested that PR is a useful feature with obvious break points close to BBCH 31 for a small study area in Germany. In comparison, a study in France by Veloso et al. (2017) suggested that PR could enhance the phenology estimates for early stages and provides insight into structural changes within the canopy. Therefore, we see our results are in line with earlier studies that proposed PR as a valuable feature set for identifying early stages. On the other hand, Gerstmann et al. (2016) used GDD and elevation data to estimate phenology for different crops in whole Germany. They reported that for early stages temperature played a major role while still with problems to estimate shooting stage in winter wheat, winter rye, winter barley, winter rapeseed, spring oat, and maize.

The middle of the season stages such as general flowering (BBCH 65) are critical agronomic stages that mark the transition from the vegetative phase to the reproductive (or regenerative) phase, which is pivotal in crop development, growth modeling, and management strategies. During this period, crops undergo essential processes such as flowering, pollination, and early fruit or grain formation, which directly influence final yield and quality and control the grain filling period. Accurate monitoring and modeling of these stages are crucial for optimizing agronomic practices, such as nutrient application, irrigation scheduling, and pest and disease management. Lobert et al. (2023) reported MAE of 4.5 days for the heading stage of winter wheat slightly less good than the MAE of 4.34 days in our study. By studying the combination of radar and optical sensors for rapeseed in eastern part of Germany, d'Andrimont et al. (2020) found that the flowering stages are detectable with temporal accuracy of 1–4 days with both radar and optical sensors, while the result of Htitiou et al. (2024) had a high MAE of 11 days for heading stage of winter wheat with using only optical sensor. Veloso et al. (2017) found that the optical sensor is more informative for the predicting the heading stage of winter wheat, while for maize, the radar sensor data was more informative. Gerstmann et al. (2016) suggested that GDD and elevation are informative for heading stage with average MAE of 2.14 days for winter wheat. Thus, we can state that the combination of different sets of features was successful with at least across the board as good and an often better performance metric (MAE) for predicting the middle stages, in comparison to literature values.

At the end of the season, when the crops are reaching their peak and delivering their maximum economic benefits, is a critical time for farmers and agricultural communities. Previous studies found that optical and climate data were more informative for predicting late stages (Gerstmann et al. 2016; Lobert et al. 2023; Meroni et al. 2021), while some found that radar sensor can be useful for cereals and mowing in grasslands (Kavats et al. 2019; Lobert et al. 2021). Radar sensors which are sensitive to surface resistance can be affected by the morphological characteristics of crops which are rather stable during late stages. For instance, since the maize height can attain 2.5 meters and more, the soil influence on the signal will then be marginal (Veloso et al. 2017) and makes it challenging to differentiate between late stages. Sugar beet in Germany, however, is affected by multiple other factors that decide upon harvest scheduling by private companies that can go as late as harvesting in February. Besides, harvested beet are normally left in field corners, which may impact negatively on signal retrieval from RS data, while residual leaves left after harvesting may introduce further noise in RS data. Most studies reported high accuracies for the end of the season in other crops which is closely related to the changes in morphology of the crops and can be detected with radar sensor and also the GDD sum and precipitation sum that identify the end of the season (Htitiou et al. 2024; Lobert et al. 2023; Schlund and Erasmi 2020; Veloso et al. 2017).

In comparison to Lobert et al. (2023), with similar approach except that they used deep learning model only for winter wheat, our accuracies and errors in $R^2$ and MAE were better for all stages (Fig. 7), especially for initial stages of BBCH 0 and 10 with improvements of $R^2$ by 0.29 and 0.35, and MAE by -1.70 and -1.87 days, respectively, compared to their results. Lobert et al. (2023) found that climate parameters have no high explanatory power and do not affect the model accuracy at field level. In contrast, the presented study shows that climate parameters are among the most important features, particularly in a heterogeneous and complex landscape like in Germany. It may be rooted in the substantial number of climate station in this study (2144 stations) compared to (Lobert et al. 2023) study (625 stations) which helped us to find local patterns of climate features. We are aligned with previous studies that suggest climate features are critical variables to detect and identify trends of phenological stages (Brown et al. 2012; Gerstmann et al. 2016; Pei et al. 2025). Similar plots like Fig. 10 for other crops can be found in the Appendix (Figs. A9-A15).

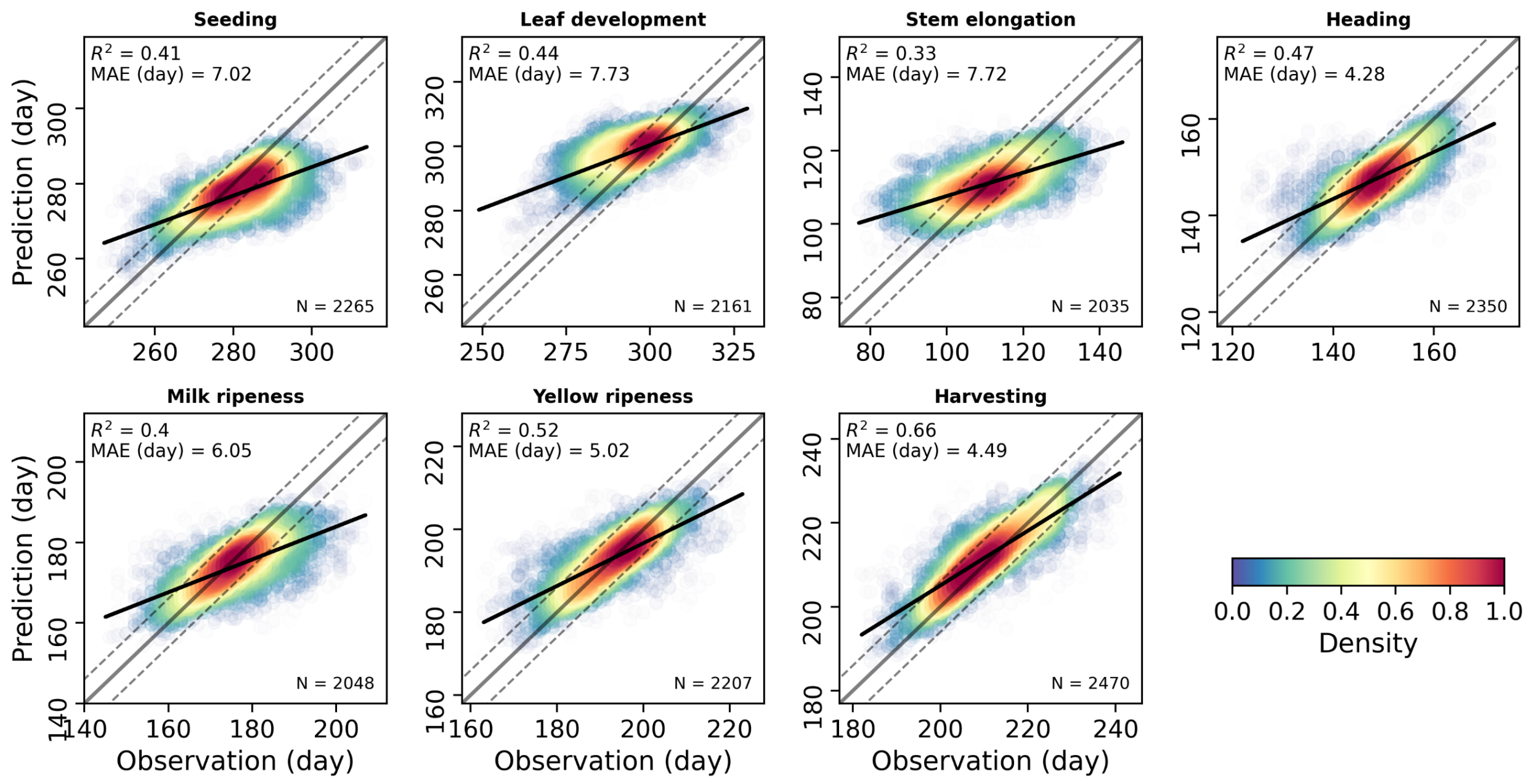

**Figure 10** Density plot of predictions versus observations for winter wheat in different phenology stage. The dashed black lines are the deviation of ±6 days from the prefect prediction.

### 4.3.Spatio-temporal evaluation

We proposed a nested random cross validation to distribute phenological stations randomly, preserving the predictive strength of the geospatial features. Therefore, the spatial transferability of the model is preserved with this method. However, the underestimations of the early stages primarily stem from the absence of cover crops during the early stages and the significant soil signal from bare soil, which is influenced by early tillage that immediate the detection of the seeding stage (Lobert et al. 2023). The reference data, which is based on volunteer-reported ground truth for DWD, provides evidence for the early stages but does not pinpoint the exact date. Consequently, it is likely that the early stage of seeding began a few days prior to the reported date. This implies that the model may perform well in slightly underestimating the seeding stage. Furthermore, the farmer may need to postpone the seeding date if the soil is too wet during heavy rainy days, potentially causing further delays (Lobert et al. 2023). For sugar beet, the underestimation for early stages may be related to different seeding dates in the same region which affect the signal of radar sensor along with tillage.

For the late stage of harvesting, even though the model overestimates the harvesting date, there is no spatial pattern similar to the one found for the seeding stage and it is affected by various reasons.

One reason has been given by Harfenmeister et al. (2021) and relates to the effect of the time series smoothing that may eliminate the break points in time series after harvesting. For instance, this may be clear for the VH time series (Fig.2). Another reason may be rooted in the post harvesting condition (or more generally different tillage operations) of the field in which after, ploughing fields are completely clear and other tillage practices may result in more residues remaining at the soil surface with different properties. Also, the radar signal is affected by the soil moisture in bare soils and some fields are covered by residues which is affecting the radar signal and has no clear break points to detect phenological stages confirmed by (Shang et al. 2020). These differences between fields may be an important source of perturbations in the data and in consequence make it more difficult for the model to detect the date of phenological stages. Sugar beet, however, with a different harvesting date in the same region or maybe in same field are affected by management practices and the averaging is comixing the radar time series which could immediate the harvest date especially for the first harvesting date (Olson et al. 2019).

The nested cross validation also randomly selects data through the years and preserves temporal transferability of the proposed method. However, patterns observed through the years and crops such as the summer crops for 2018 are consistent with those of the winter crops which have a broader window of growth through the whole year. This may give winter crops some resistance to short-term changes in specific weather conditions such as warm spring or heat waves in summer, as the crops are not entirely reliant on a specific season or time window for the development. While winter crops seem more reliable in terms of spatial and temporal transferability of the model due to their longer growth cycles, they are not immune to extreme climate events (Estrella et al. 2007). Management practices such as changing cultivar through the years, altering sowing density, and time of sowing selection of heat-tolerant varieties, or water management practices according to shifting environmental conditions (Rezaei et al. 2017; Rezaei et al. 2018), may be confounding the ML-based prediction algorithms. In contrast, summer crops, which are usually planted during spring and harvested during the summer season, are more exposed to the changing time of seasons and extremes in climate, such as heatwaves and droughts (Chmielewski et al. 2004). These crops become increasingly vulnerable during their shortened growth cycle to even slight changes in the pattern of planting dates or excessive temperature anomalies. Therefore, winter crops in terms of temporal transferability of the model are more reliant compared to summer crops.

### 4.4. Study limitations and outlook

We shaped our study based on DWD observations collected by trained volunteers across Germany. Like any observation, aleatoric uncertainties (here human errors) occur during the data gathering process (here, field observation) are an inevitable part (Liu et al. 2021a). For instance, some of the sampling points had insufficient location precision (locations with 1 decimal precision; for example, 4.4 as a longitude of a DWD station, which is not precise enough for pixel accuracy of 20 m in this study). Therefore, not only the sampling uncertainty affecting the reference data (known as human error) but also sampling strategy are not established/reported observation points with required precision in some cases. Thus, we encourage improving the accuracy in the reporting of the geolocation of DWD stations.

On the other hand, the pixel size issues (known as footprint mismatch between observations and RS data) in RS application is always a common issue which is probable to have affected modeling accuracy and imposes undesirable uncertainties, especially when focusing on time series modeling (Povey and Grainger 2015). Also, footprint mismatch can affect the boundaries of different crop fields, which is substantially changes the map of phenological developments like its effects on land cover mapping (Lechner et al. 2009). This footprint mismatch also is prevalent in CTM data.

The biases and uncertainties in RS data are not solely related to pixel size. Various factors such as climate condition, cloud coverage, accuracy of cloud removal, and especially the noise in time series modeling based on RS data could relatively affect the modeling precision (Li et al. 2022b; Zhou et al. 2016). Additionally, the noise reduction procedure is always affecting the peak values of agricultural indices (e.g., PR) and could introduce uncertainties to the modeling process. This point can affect the precision (Shao et al. 2016).

The density of climate stations near DWD stations can influence the accuracy of capturing the variability of climate patterns at both local and regional levels (Gerstmann et al. 2016). Figure A1 illustrates that while some stations are surrounded by numerous climate stations that can aid in identifying microclimate factors, others lack sufficient nearby stations. This scarcity may introduce uncertainty when using IDW to determine regional and local microclimate patterns (Tomczak 1998).

ML-based models rely on extensive phenological ground observations to be effectively integrated with RS data. However, collecting such ground-based phenological data can be challenging in many countries. A possible avenue to tackle this challenge could be utilizing big datasets such as Pan European Phenology (PEP) project (Templ et al. 2018) to construct an extensive ML model in future studies. Moreover, ML models function as black boxes, obscuring the processes behind predictions and potentially limiting our ability to understand how factors lead to detection of phenological stages.

Within-season crop phenology detection remains an open challenge (Gao and Zhang 2021). Phenological methods such as proposed method generally require prior knowledge of crop type to detect phenological stages (Cao et al. 2024); however, acquiring this information within-season on a national scale is challenging. Even in the best-case scenario, crop type maps may only become available after the current season, with delays frequently extending even further (Blickensdörfer et al. 2022). Moreover, phenology prediction often depends on post-phenological stage data (RS data) to accurately detect the stage, even in methods that do not require crop type information, making it a persistent challenge for within-season phenology predictions (Gao et al. 2020).

Future studies also could explore the incorporation of automated drones and unmanned aerial vehicle (UAV) at large farm and regional scales to estimate crop phenology in conjunction with using radar, optical, and climate data (Maurya et al. 2023). Advanced imaging technologies can be employed by drones to improve the efficiency and accuracy of data gathering on real-time monitoring of crop health and stages of growth (Lu et al. 2023). This approach not only stands in support of the digitalization of farming practices but also aligns with our current methods and may provide scalable applications in precision agriculture. The investigation of the interaction between drone technology and other available data sources may provide substantial added value both on crop management and sustainability.

## 5. Summary and conclusions

In this study we evaluated the fusion of optical, radar, and climate data integrated with an ML model to detect phenological development of eight crops and different plant growth stages across Germany. We used diverse RS and climate indices along with raw bands and backscatter coefficient data to explore the efficiency of combining multiple RS and climate parameters on the precision of a proposed ML model. We further improved the ML model accuracy using a hyper tuning approach and proposing multiple feature set based on satellites and climate parameters to understand underlying characteristics of crop growth.

Satellite and climate parameters along with phenological observations of DWD stations were arranged to create a training dataset to build and calibrate an ML model with the aim to detect phenological stages of crops and predict the date of each stage. To evaluate the ML model's accuracy, we did a nested cross validation with outer and inner loops splitting data to 10 folds. In the inner loops we hyper tuned the ML model along with selecting the best features with 50 trials. Then the most repeated features are selected to retrain the ML model. We found there were no significant changes in the model's accuracy. Then, the model's accuracy was evaluated resulting in an average MAE < 5.83 (days) and $R^2 > 0.42$ over all crops and BBCH stages. In tendency, better values were achieved during the mid-season, than during the early stage of the season (BBCH 31~shooting). Additionally, the proposed model could effectively find spatio-temporal variability and patterns and was very transferable through space and time for winter crops across Germany.

We summarize our findings as:

1- Hyper-tuning of ML model is a necessary step to find and fairly compare the potential of each data source (RS and/or climate) in detecting phenological stages.
2- The early stages of crops are challenging to detect. Spatial features such as latitude and longitude, elevation and its derivatives can help inform the model about these differences.
3- RS and climate data together are essential for effectively detecting phenological stages; neither can do it alone.
4- Management practices are critical factors that make it difficult to effectively detect phenological stages and lead to overestimation in the late stages and underestimation in early stages.
5- Winter crops show stable model performance, while summer crops exhibit greater variability due to extreme weather, emphasizing the need for adaptive seasonal modeling.
6- The proposed method effectively identifies phenological stages, particularly the crucial transition from heading to seeding emergence. This aids crop modelers in enhancing crop growth models and supports farmers by providing insights for key decisions, such as irrigation and fertilization, through the digitalization of agriculture.

## Funding

This research was funded by the joint project of Digitalization in organic agriculture (DigiPlus, grant no. 28 DE 207 A 21).

## Data availability

All satellite data sources used in this study are available freely from Google Earth Engine data catalogue, DWD observations through (https://opendata.dwd.de/climate_environment/CDC/observations_germany/phenology/),

CTM raster data through (https://zenodo.org/records/10617623), and climate data using Meteostat Python package (https://dev.meteostat.net/python/).

## Appendix

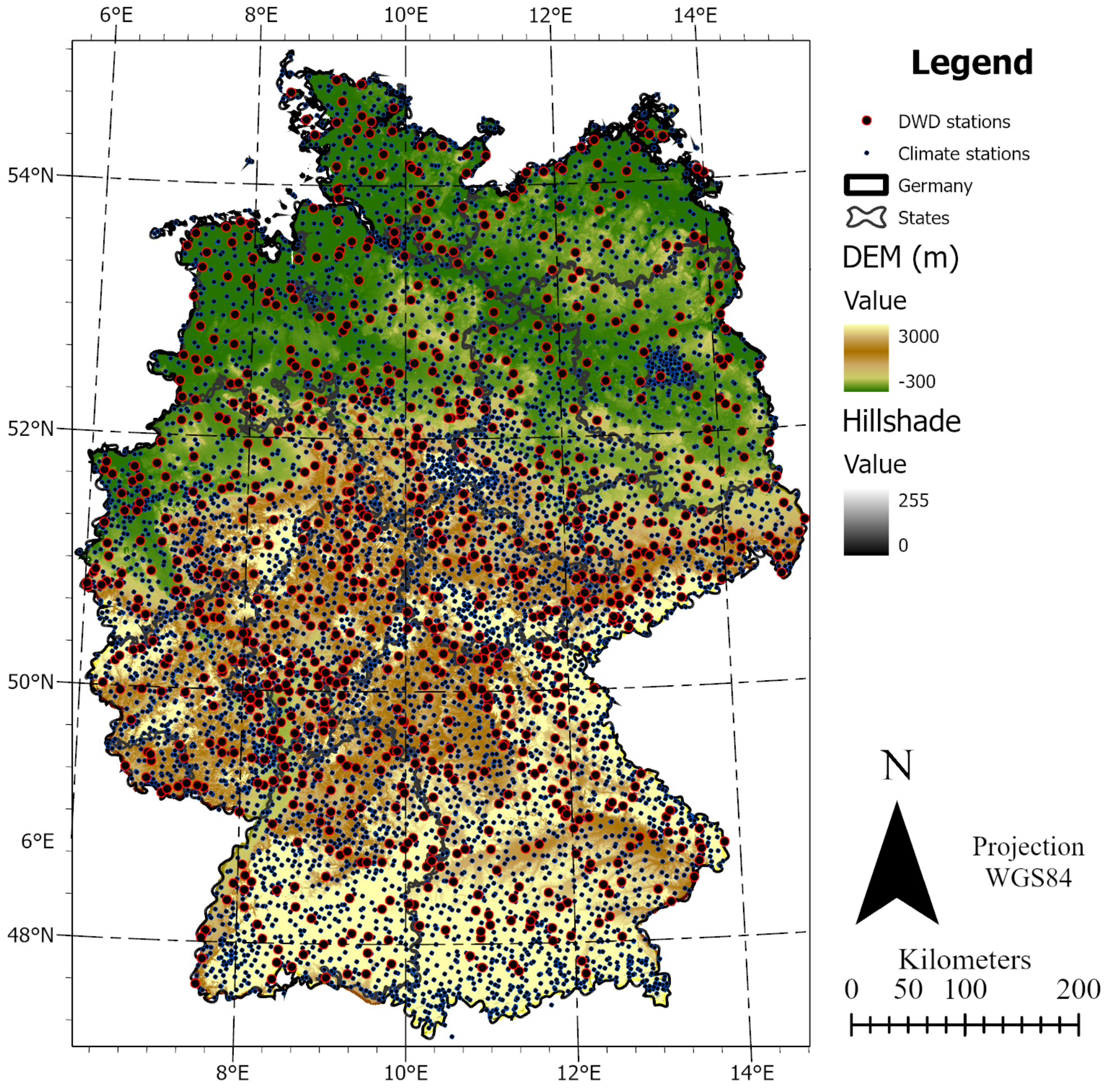

**Figure A1** Distribution of DWD and climate stations across Germany.

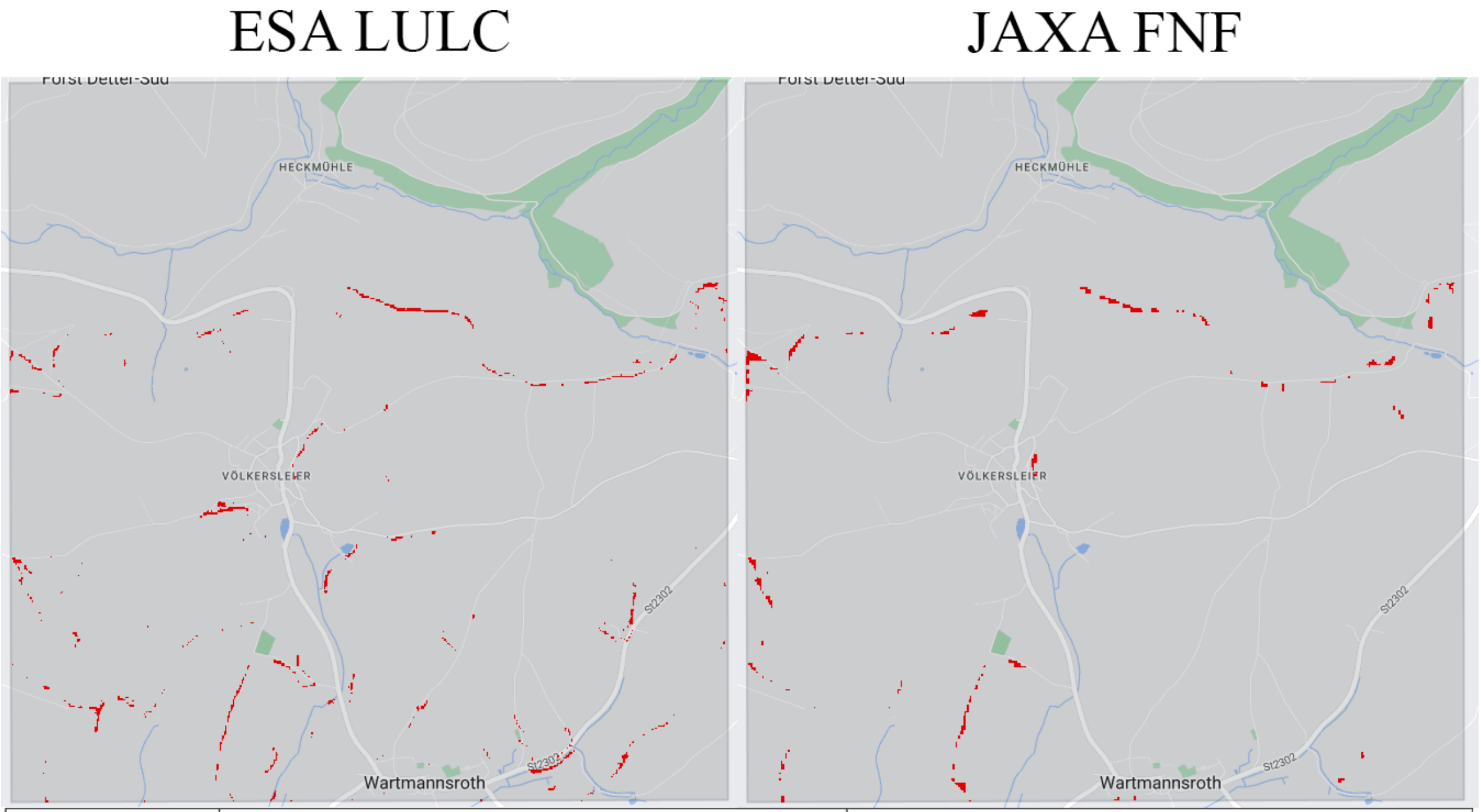

| Crop | ESA LULC (10 m) | | | | JAXA FNF (25 m) | | | |
|---|---|---|---|---|---|---|---|---|
| | Area (Ha) | Diff (mean) | Diff (min) | Diff (max) | Area (Ha) | Diff (mean) | Diff (min) | Diff (max) |
| Winter wheat | 1049 | 4.4 | -8 | 19 | 969 | 7.5 | 0.6 | 19 |
| Winter barley | 962 | 6.6 | -0.5 | 27.2 | 653 | 11.6 | 0.3 | 27.2 |
| Winter rye | 619 | 5.6 | -10 | 14.7 | 579 | 8.8 | 0.7 | 14.7 |
| Spring barley | 249 | 5.9 | -23.6 | 11.9 | 332 | 3.4 | -23.6 | 11.9 |
| Spring oat | 192 | 7.7 | -4.6 | 22.3 | 198 | 10.9 | 4.3 | 22.3 |
| Maize | 2378 | 6.7 | -0.3 | 16.7 | 3781 | 7.4 | -1.2 | 32.9 |
| Sugar beet | 217 | 9.2 | -0.2 | 35 | 309 | 5.3 | -0.2 | 11.3 |
| Winter rapeseed | 599 | 4.7 | -3.2 | 20 | 995 | 7.7 | -0.2 | 20 |

Note: Diff is the difference between Copernicus DEM and FABDEM in meters.

**Figure A2** Red pixels show areas that are classified as a crop type while they are classified as forest (left) in ESA Land Use Land Cover (LULC) and (right) in JAXA Forest/Non-Forest (FNF) classification. The table at the bottom represents the statistics of difference between (in meters) CDEM and FABDEM in all pixels for each crop type in all 862 DWD stations. Comparison of CTM data with ESA World Cover (Zanaga et al. 2021) and JAXA forest/non-forest data (Shimada et al. 2014) for 2020, assessing misclassified crop boundaries and differences between FABDEM and CDEM at 862 DWD stations for eight crops. Results support the use of FABDEM for spatial phenology estimation in agroforestry regions like Germany.

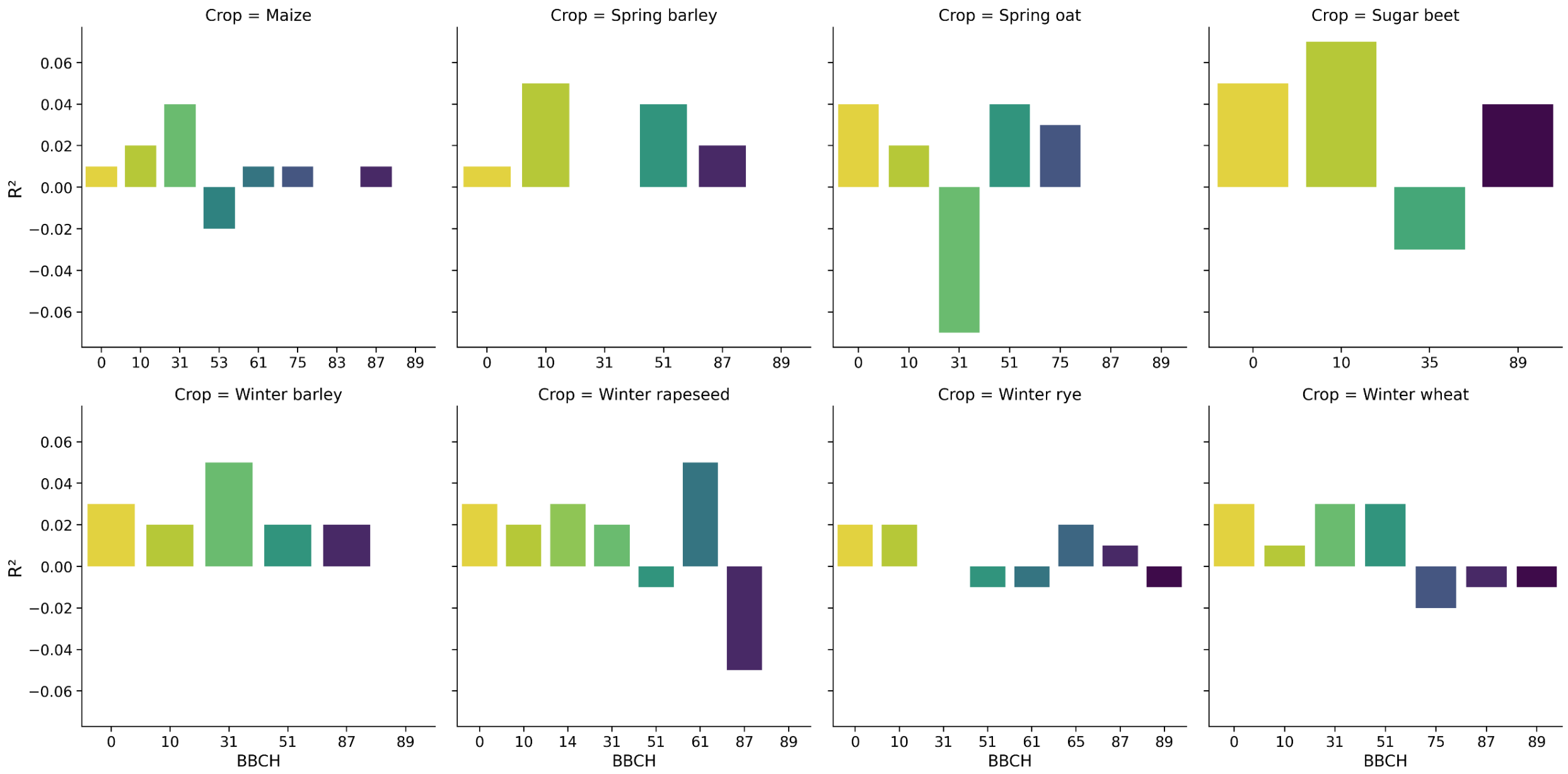

**Figure A3** The difference between $R^2$ of the standardized feature sets and best feature set for each crop and each BBCH.

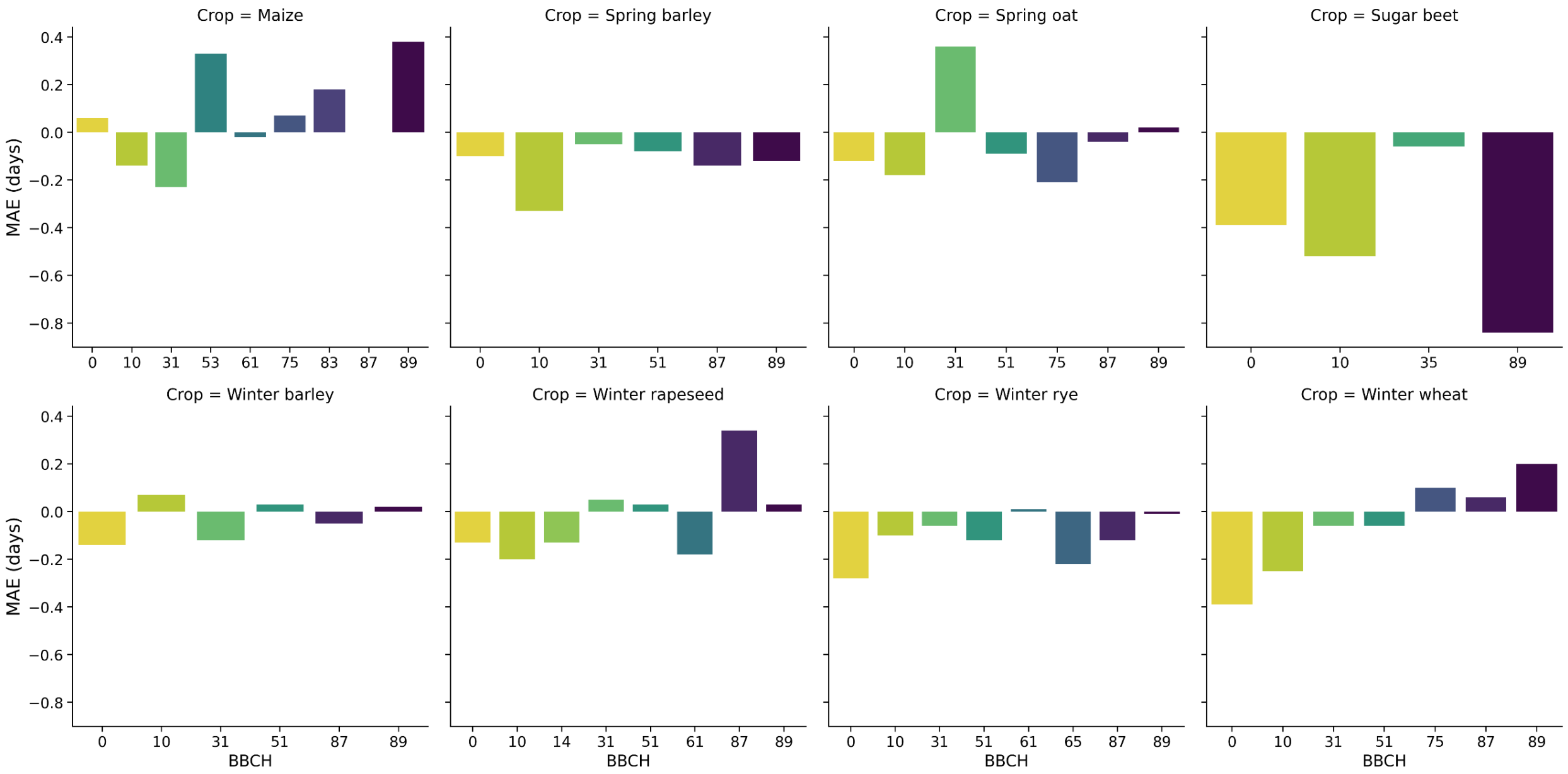

**Figure A4** The difference between MAE of the standardized feature sets and best feature set for each crop and each BBCH.

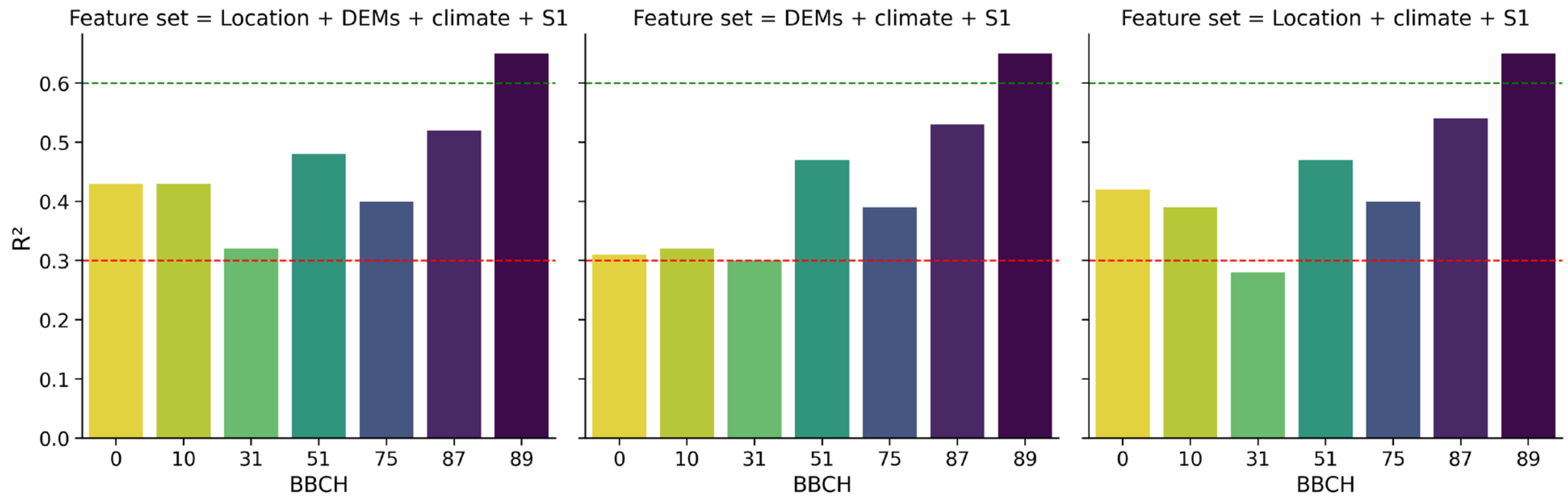

**Figure A5** The difference between $R^2$ of the standardized feature sets and removing the static features for winter wheat.

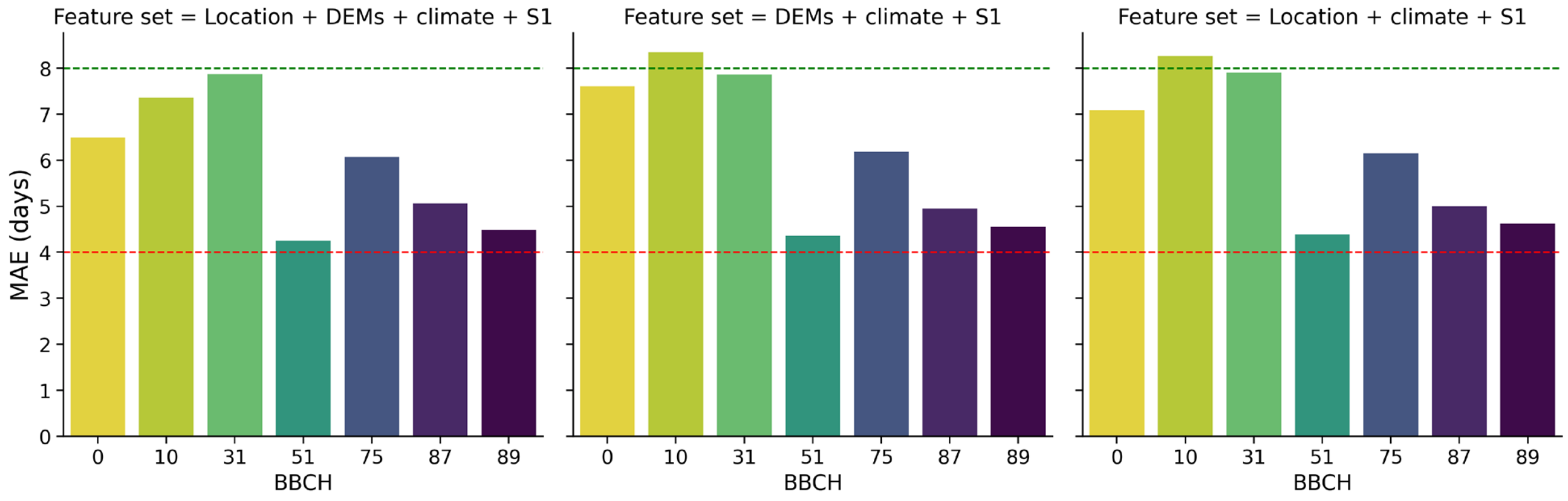

**Figure A6** The difference between MAE of the standardized feature sets and removing the static features for winter wheat.

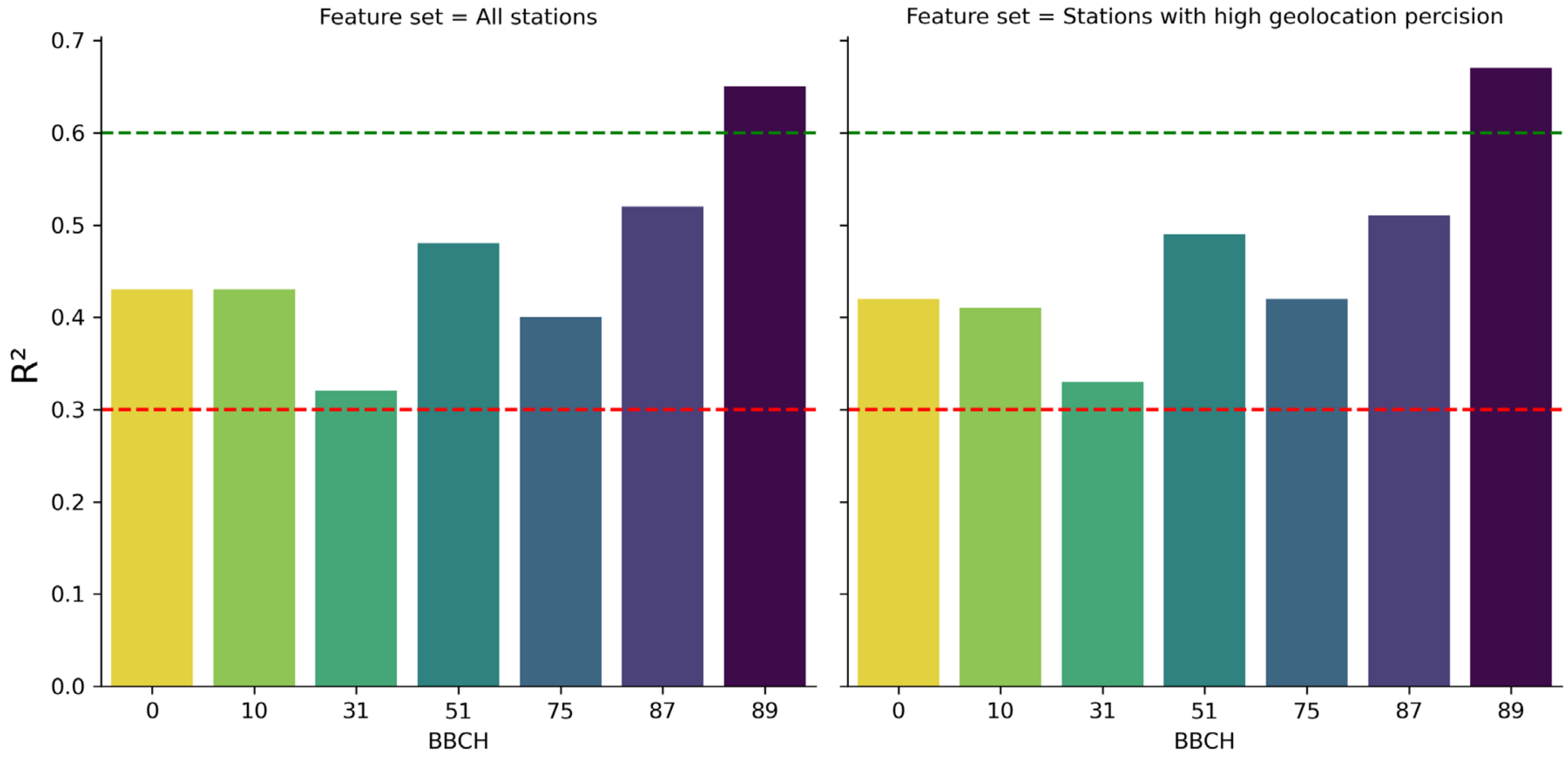

**Figure A7** The difference between $R^2$ of all stations (including low-precision ones) and high-precision geolocation stations for winter wheat.

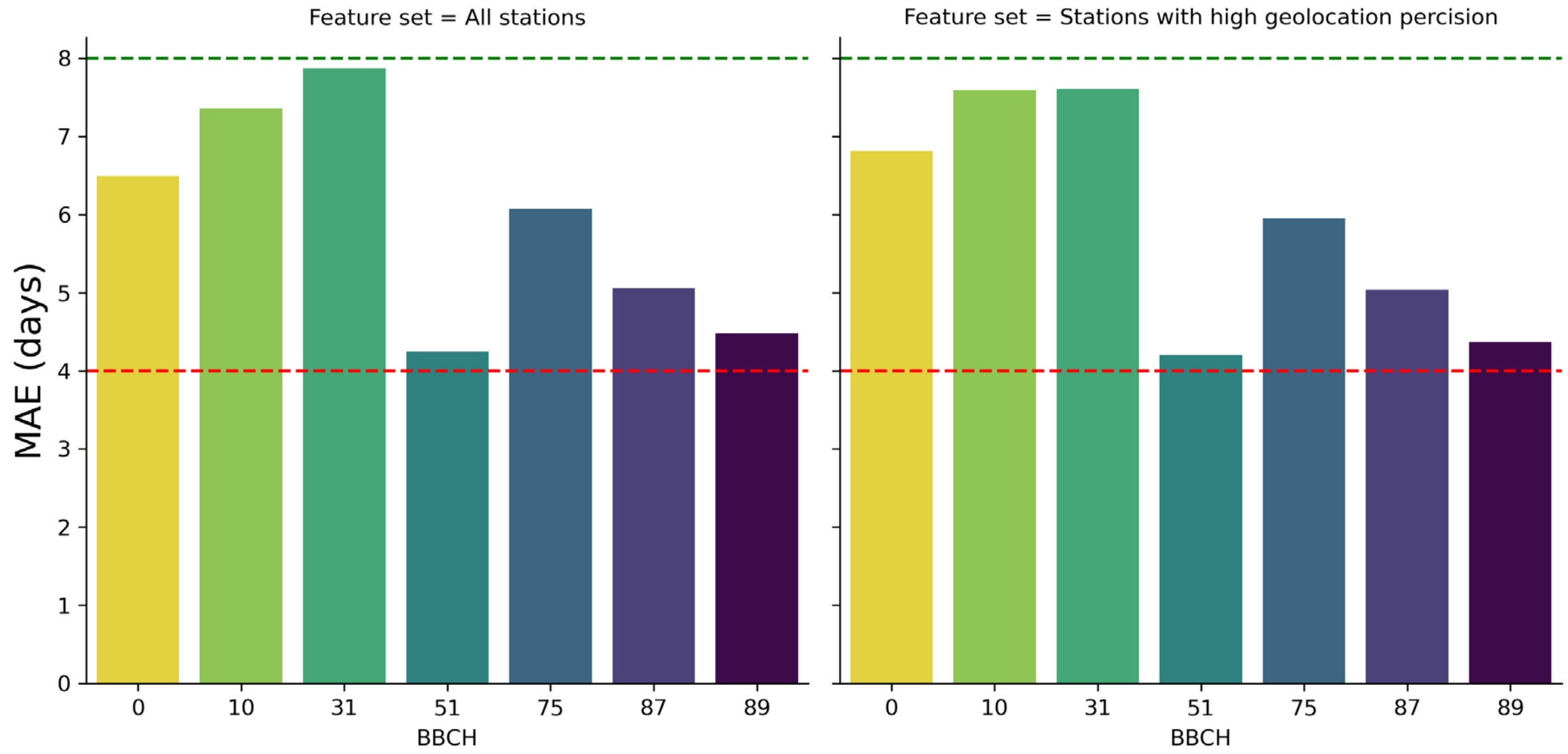

**Figure A8** The difference between MAE of all stations (including low-precision ones) and high-precision geolocation stations for winter wheat.

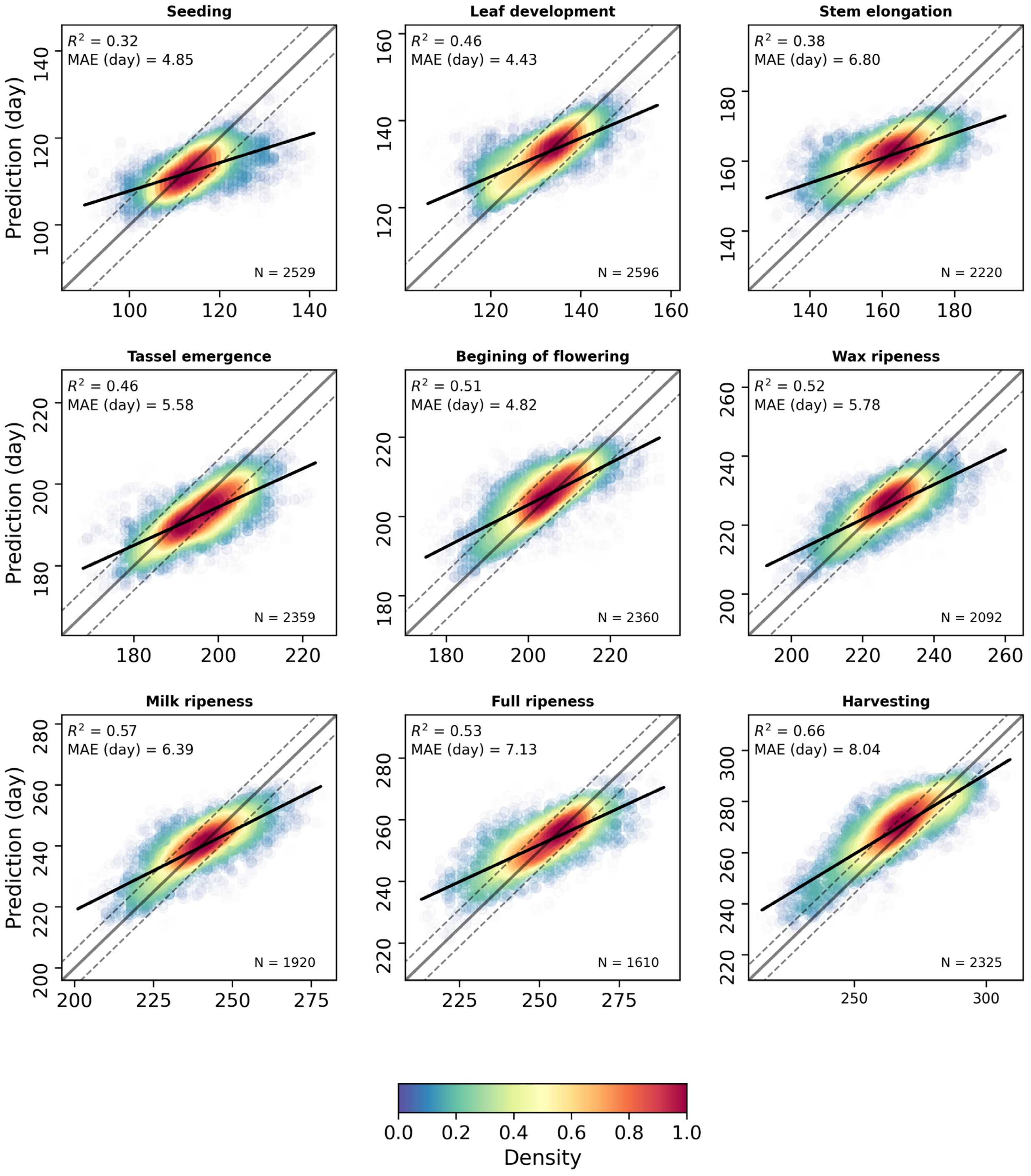

**Figure A9** Density plot of predictions versus observations for maize in different phenology stage. The dashed black lines are the deviation of ±6 days from the prefect prediction.

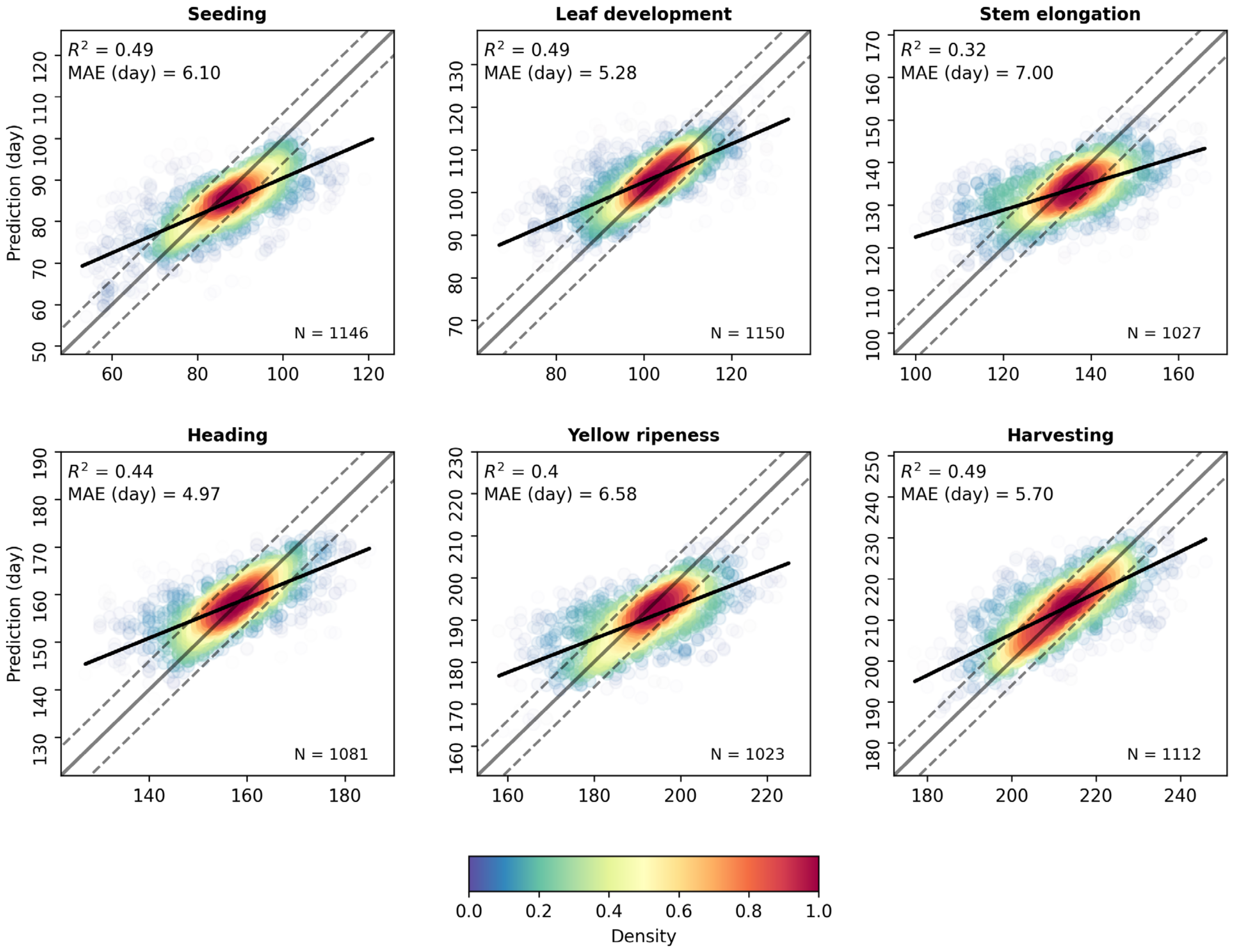

**Figure A10** Density plot of predictions versus observations for spring barley in different phenology stage. The dashed black lines are the deviation of ±6 days from the prefect prediction.

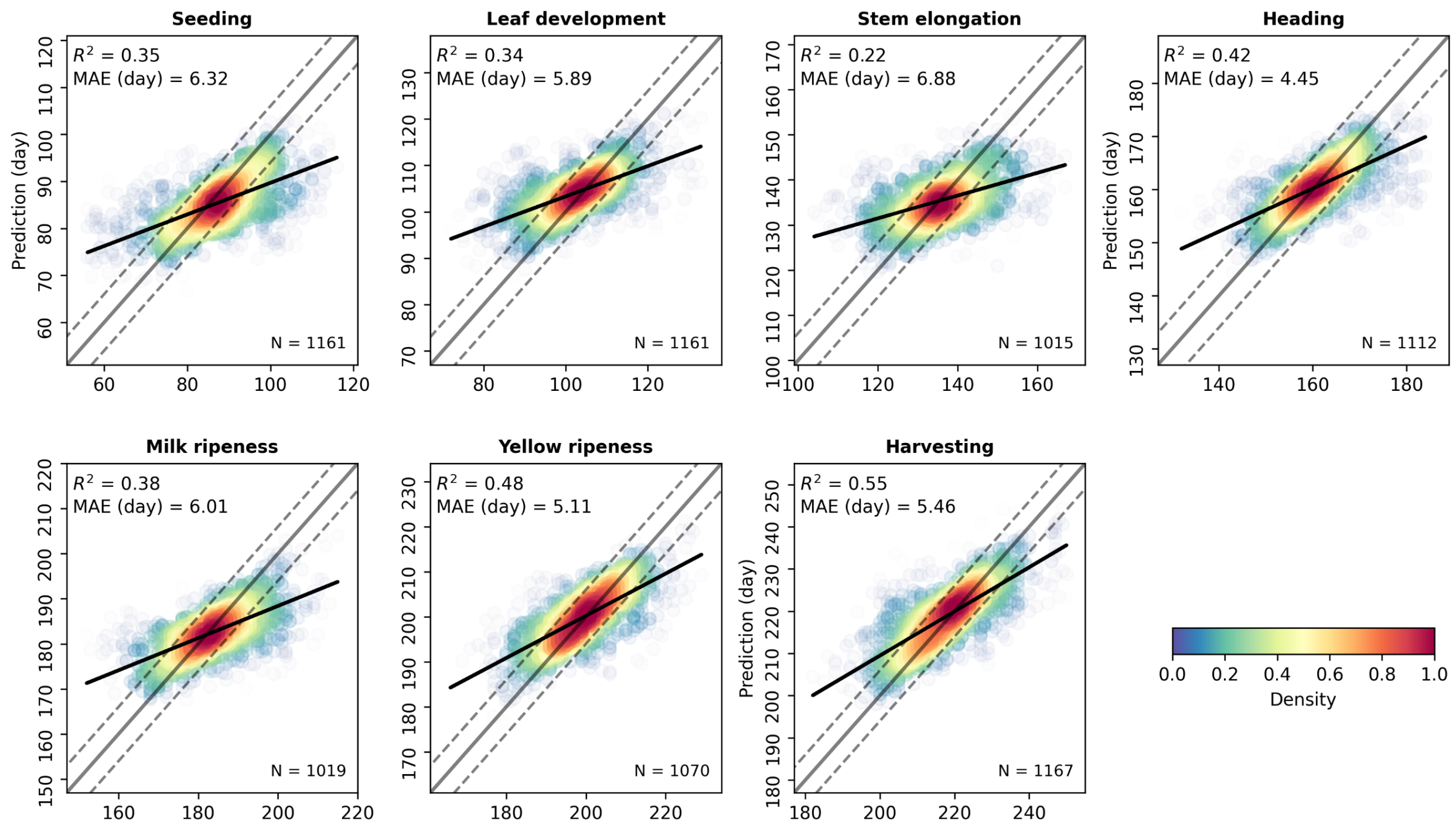

**Figure A11** Density plot of predictions versus observations for spring oat in different phenology stage. The dashed black lines are the deviation of ±6 days from the prefect prediction.

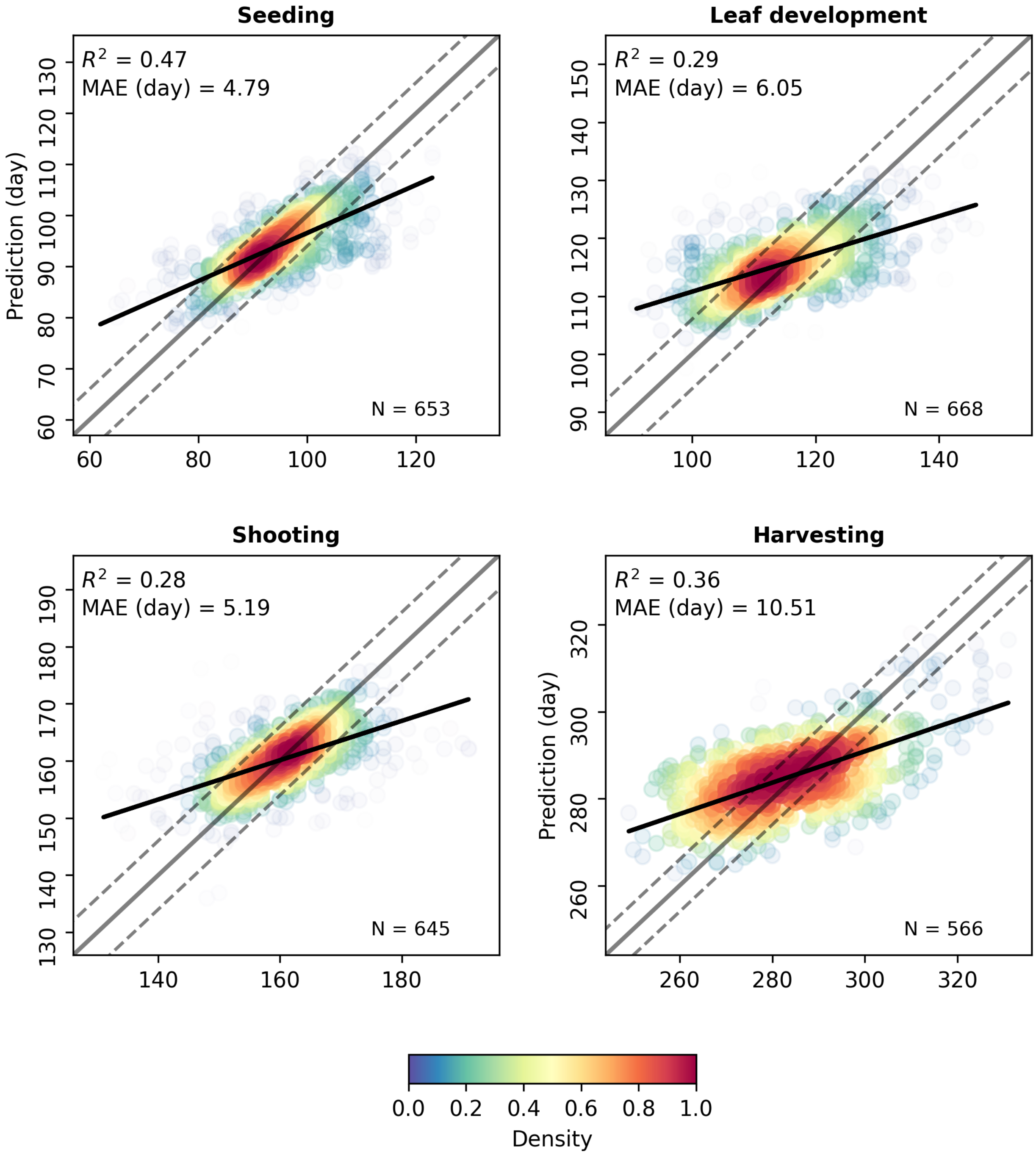

**Figure A12** Density plot of predictions versus observations for sugar beet in different phenology stage. The dashed black lines are the deviation of ±6 days from the prefect prediction.

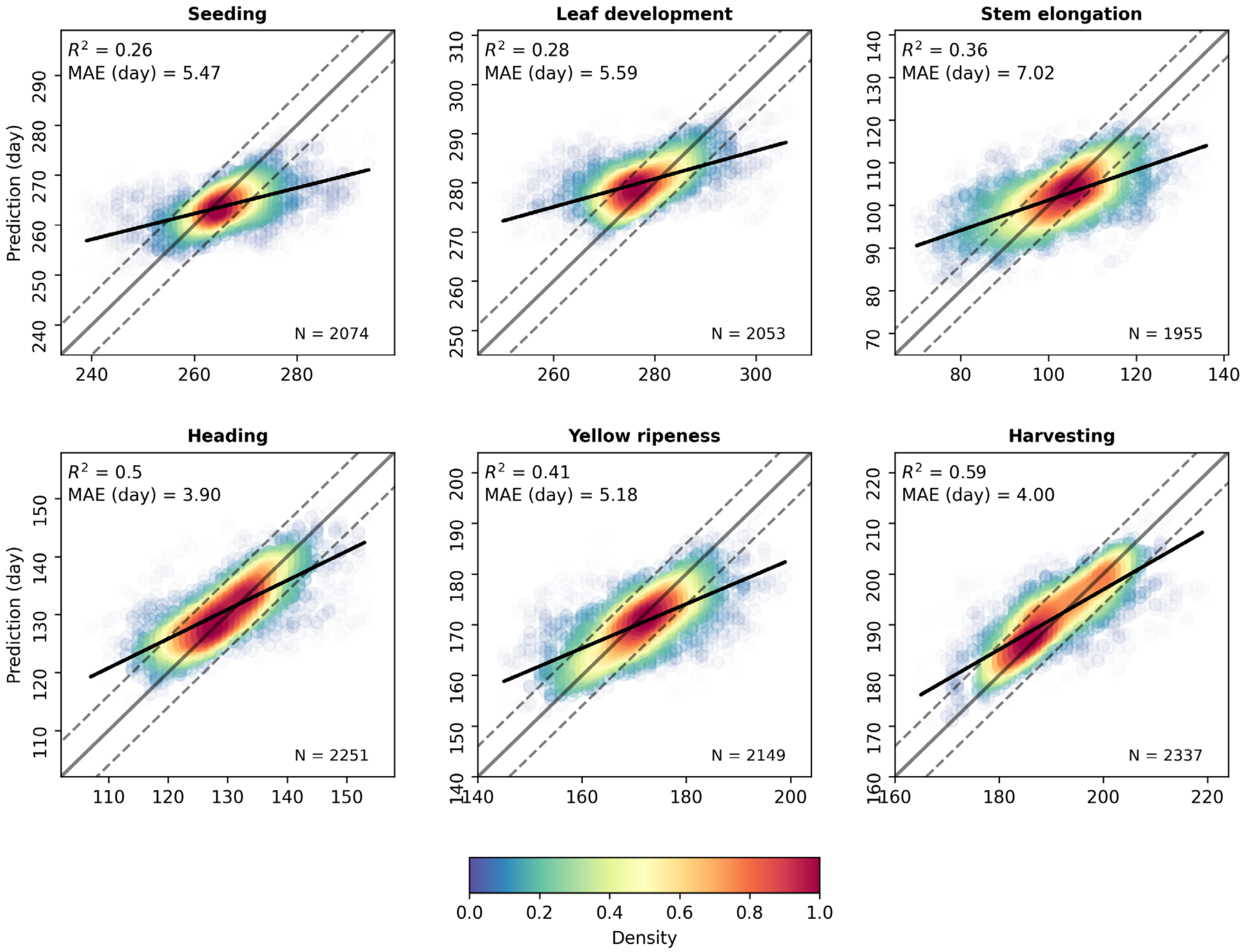

**Figure A13** Density plot of predictions versus observations for winter barley in different phenology stage. The dashed black lines are the deviation of ±6 days from the prefect prediction.

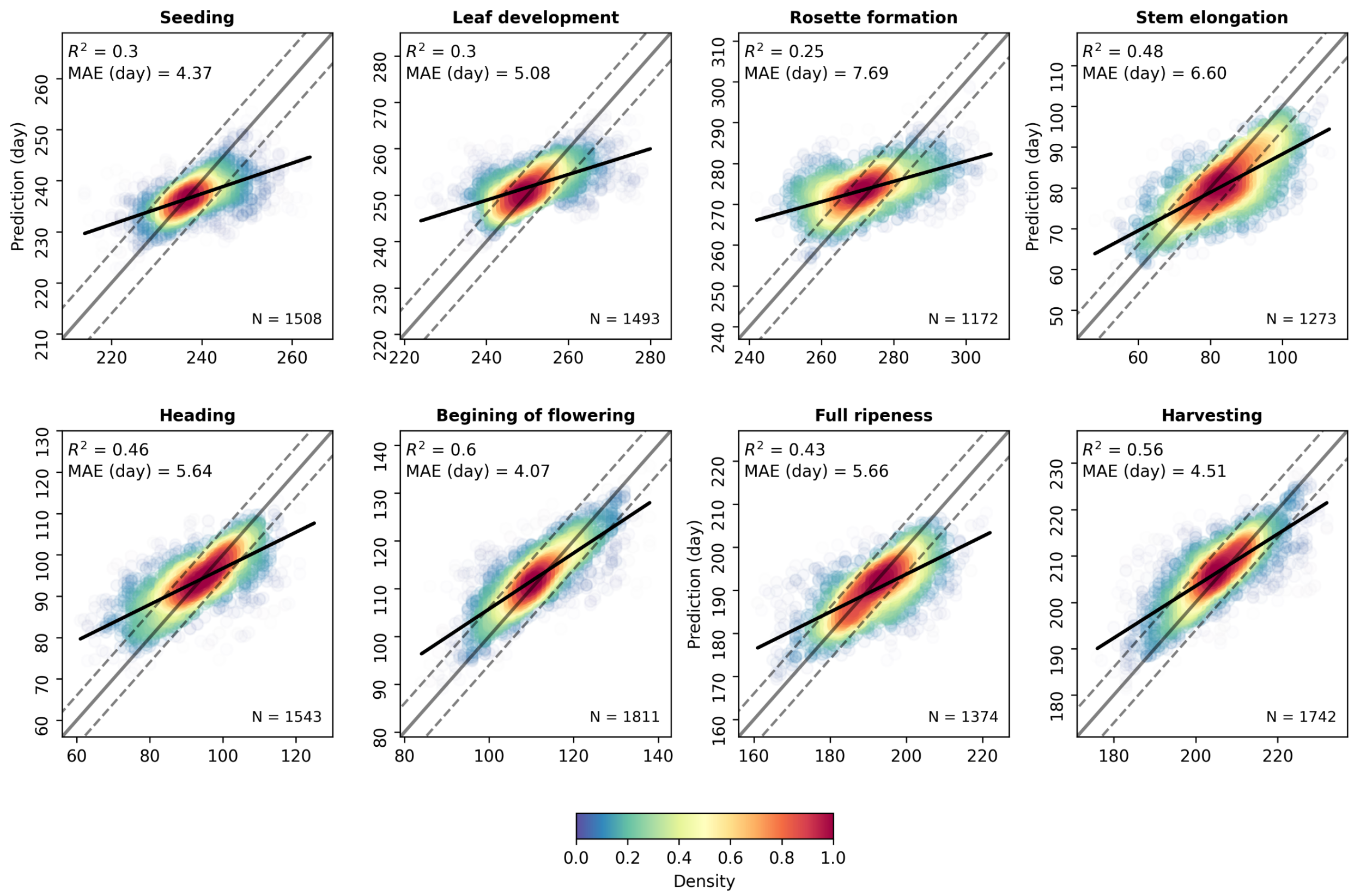

**Figure A14** Density plot of predictions versus observations for winter rapeseed in different phenology stage. The dashed black lines are the deviation of ±6 days from the prefect prediction.

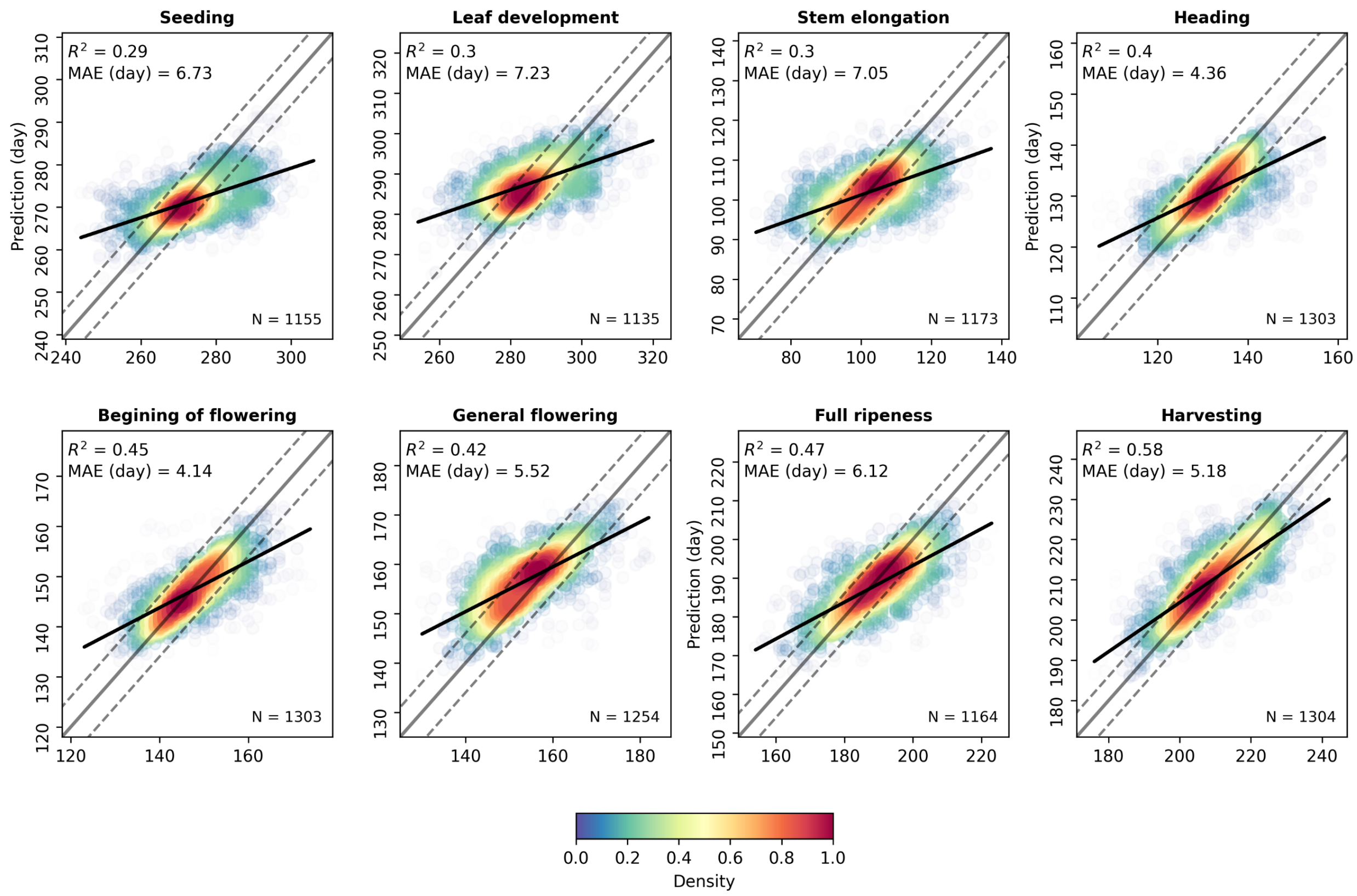

**Figure A15** Density plot of predictions versus observations for winter rye in different phenology stage. The dashed black lines are the deviation of ±6 days from the prefect prediction.

## References

Akiba, T., Sano, S., Yanase, T., Ohta, T., & Koyama, M. (2019). Optuna: A Next-generation Hyperparameter Optimization Framework. In, *Proceedings of the 25th ACM SIGKDD International Conference on Knowledge Discovery & Data Mining* (pp. 2623–2631). Anchorage, AK, USA: Association for Computing Machinery

Aslam, M.A., Ahmed, M., Stöckle, C.O., Higgins, S.S., Hassan, F.u., & Hayat, R. (2017). Can growing degree days and photoperiod predict spring wheat phenology? *Frontiers in Environmental Science, 5*, 57

Babcock, C., Finley, A.O., & Looker, N. (2021). A Bayesian model to estimate land surface phenology parameters with harmonized Landsat 8 and Sentinel-2 images. *Remote Sensing of Environment, 261*, 112471

Badeck, F.W., Bondeau, A., Böttcher, K., Doktor, D., Lucht, W., Schaber, J., & Sitch, S. (2004). Responses of spring phenology to climate change. *New Phytologist, 162*, 295-309

Baroni, G., Drastig, K., Lichtenfeld, A.-U., Jost, L., & Claas, P. (2019). Assessment of Irrigation Scheduling Systems in Germany: Survey of the Users and Comparative Study. *Irrigation and Drainage, 68*, 520-530

Beleites, C., Neugebauer, U., Bocklitz, T., Krafft, C., & Popp, J. (2013). Sample size planning for classification models. *Analytica Chimica Acta, 760*, 25-33

Blickensdörfer, L., Schwieder, M., Pflugmacher, D., Nendel, C., Erasmi, S., & Hostert, P. (2022). Mapping of crop types and crop sequences with combined time series of Sentinel-1, Sentinel-2 and Landsat 8 data for Germany. *Remote Sensing of Environment, 269*, 112831

Brown, M.E., de Beurs, K.M., & Marshall, M. (2012). Global phenological response to climate change in crop areas using satellite remote sensing of vegetation, humidity and temperature over 26years. *Remote Sensing of Environment, 126*, 174-183

Canisius, F., Shang, J.L., Liu, J.G., Huang, X.D., Ma, B.L., Jiao, X.F., Geng, X.Y., Kovacs, J.M., & Walters, D. (2018). Tracking crop phenological development using multi-temporal polarimetric Radarsat-2 data. *Remote Sensing of Environment, 210*, 508-518

Cao, R., Li, L., Liu, L., Liang, H., Zhu, X., Shen, M., Zhou, J., Li, Y., & Chen, J. (2024). A spatiotemporal shape model fitting method for within-season crop phenology detection. *ISPRS Journal of Photogrammetry and Remote Sensing, 217*, 179-198

Center, M.R.C. (2001). Midwestern Regional Climate Center. *ISWS Informational/Educational Materials 2001-01*

Chen, T., & Guestrin, C. (2016). Xgboost: A scalable tree boosting system. In, *Proceedings of the 22nd acm sigkdd international conference on knowledge discovery and data mining* (pp. 785-794)

Chmielewski, F.-M., Müller, A., & Bruns, E. (2004). Climate changes and trends in phenology of fruit trees and field crops in Germany, 1961–2000. *Agricultural and Forest Meteorology, 121*, 69-78

Courter, J.R., Johnson, R.J., Stuyck, C.M., Lang, B.A., & Kaiser, E.W. (2013). Weekend bias in Citizen Science data reporting: implications for phenology studies. *International Journal of Biometeorology, 57*, 715-720

Czernecki, B., Nowosad, J., & Jablonska, K. (2018). Machine learning modeling of plant phenology based on coupling satellite and gridded meteorological dataset. *Int J Biometeorol, 62*, 1297-1309

d'Andrimont, R., Taymans, M., Lemoine, G., Ceglar, A., Yordanov, M., & van der Velde, M. (2020). Detecting flowering phenology in oil seed rape parcels with Sentinel-1 and -2 time series. *Remote Sensing of Environment, 239*, 111660

Dandabathula, G., Hari, R., Ghosh, K., Bera, A.K., & Srivastav, S.K. (2023). Accuracy assessment of digital bare-earth model using ICESat-2 photons: analysis of the FABDEM. *Modeling Earth Systems and Environment, 9*, 2677-2694

De Bernardis, C., Vicente-Guijalba, F., Martinez-Marin, T., & Lopez-Sanchez, J.M. (2016). Contribution to real-time estimation of crop phenological states in a dynamical framework based on NDVI time series: Data fusion with SAR and temperature. *IEEE Journal of Selected Topics in Applied Earth Observations and Remote Sensing, 9*, 3512-3523

Delegido, J., Verrelst, J., Alonso, L., & Moreno, J. (2011). Evaluation of Sentinel-2 Red-Edge Bands for Empirical Estimation of Green LAI and Chlorophyll Content. *Sensors, 11*, 7063-7081

Diao, C.Y., Yang, Z.J., Gao, F., Zhang, X.Y., & Yang, Z.W. (2021). Hybrid phenology matching model for robust crop phenological retrieval. *ISPRS Journal of Photogrammetry and Remote Sensing, 181*, 308-326

Estrella, N., Sparks, T.H., & Menzel, A. (2007). Trends and temperature response in the phenology of crops in Germany. *Global Change Biology, 13*, 1737-1747

Eyshi Rezaei, E., Siebert, S., & Ewert, F. (2017). Climate and management interaction cause diverse crop phenology trends. *Agricultural and Forest Meteorology, 233*, 55-70

Falkner, S., Klein, A., & Hutter, F. (2018). BOHB: Robust and efficient hyperparameter optimization at scale. In, *International conference on machine learning* (pp. 1437-1446): PMLR

Feng, Z., Cheng, Z., Ren, L., Liu, B., Zhang, C., Zhao, D., Sun, H., Feng, H., Long, H., Xu, B., Yang, H., Song, X., Ma, X., Yang, G., & Zhao, C. (2024). Real-time monitoring of maize phenology with the VI-RGS composite index using time-series UAV remote sensing images and meteorological data. *Computers and Electronics in Agriculture, 224*, 109212

Fu, Y., Zhang, H., Dong, W., & Yuan, W. (2014). Comparison of phenology models for predicting the onset of growing season over the Northern Hemisphere. *PLoS ONE, 9*, e109544

Gao, F., Anderson, M., Daughtry, C., Karnieli, A., Hively, D., & Kustas, W. (2020). A within-season approach for detecting early growth stages in corn and soybean using high temporal and spatial resolution imagery. *Remote Sensing of Environment, 242*, 111752

Gao, F., & Zhang, X. (2021). Mapping crop phenology in near real-time using satellite remote sensing: Challenges and opportunities. *Journal of Remote Sensing*

Gao, Y., Wang, L.G., Zhong, G.J., Wang, Y.T., & Yang, J.H. (2023). Potential of Remote Sensing Images for Soil Moisture Retrieving Using Ensemble Learning Methods in Vegetation-Covered Area. *IEEE Journal of Selected Topics in Applied Earth Observations and Remote Sensing, 16*, 8149-8165

Gerstmann, H., Doktor, D., Glässer, C., & Möller, M. (2016). PHASE: A geostatistical model for the Kriging-based spatial prediction of crop phenology using public phenological and climatological observations. *Computers and Electronics in Agriculture, 127*, 726-738

Guth, P.L., & Geoffroy, T.M. (2021). LiDAR point cloud and ICESat-2 evaluation of 1 second global digital elevation models: Copernicus wins. *Transactions in GIS, 25*, 2245-2261

Haldar, D., Verma, A., Kumar, S., & Chauhan, P. (2021). Estimation of mustard and wheat phenology using multi-date Shannon entropy and Radar Vegetation Index from polarimetric Sentinel- 1. *Geocarto International, 37*, 5935-5962

Haldar, D., Verma, A., Kumar, S., & Chauhan, P. (2022). Estimation of mustard and wheat phenology using multi-date Shannon entropy and Radar Vegetation Index from polarimetric Sentinel-1. *Geocarto International, 37*, 5935-5962

Harfenmeister, K., Itzerott, S., Weltzien, C., & Spengler, D. (2021). Detecting phenological development of winter wheat and winter barley using time series of sentinel-1 and sentinel-2. *Remote Sensing, 13*, 5036

Hatfield, J.L., & Prueger, J.H. (2015). Temperature extremes: Effect on plant growth and development. *Weather and Climate Extremes, 10*, 4-10

Hawker, L., Uhe, P., Paulo, L., Sosa, J., Savage, J., Sampson, C., & Neal, J. (2022). A 30 m global map of elevation with forests and buildings removed. *Environmental Research Letters, 17*, 024016

Holen, C., & Dexter, A. (1996). A growing degree day equation for early sugarbeet leaf stages. *Res Ext Rep, 27*, 152-157

Htitiou, A., Möller, M., Riedel, T., Beyer, F., & Gerighausen, H. (2024). Towards Optimising the Derivation of Phenological Phases of Different Crop Types over Germany Using Satellite Image Time Series. *Remote Sensing, 16*, 3183

Hu, A., Nie, Y., Yu, G., Han, C., He, J., He, N., Liu, S., Deng, J., Shen, W., & Zhang, G. (2019). Diurnal temperature variation and plants drive latitudinal patterns in seasonal dynamics of soil microbial community. *Frontiers in Microbiology, 10*, 674

Hu, X., Li, L., Huang, J., Zeng, Y., Zhang, S., Su, Y., Hong, Y., & Hong, Z. (2024). Radar vegetation indices for monitoring surface vegetation: Developments, challenges, and trends. *Science of The Total Environment*, 173974

Huang, Y., Jiang, N., Shen, M., & Guo, L. (2020). Effect of preseason diurnal temperature range on the start of vegetation growing season in the Northern Hemisphere. *Ecological Indicators, 112*, 106161

Huete, A., Didan, K., Miura, T., Rodriguez, E.P., Gao, X., & Ferreira, L.G. (2002). Overview of the radiometric and biophysical performance of the MODIS vegetation indices. *Remote Sensing of Environment, 83*, 195-213

Jackson, M.T. (1966). Effects of microclimate on spring flowering phenology. *Ecology, 47*, 407-415

Kang, Y., Meng, Q., Liu, M., Zou, Y., & Wang, X. (2021). Crop Classification Based on Red Edge Features Analysis of GF-6 WFV Data. *Sensors (Basel), 21*

Kaspar, F., Zimmermann, K., & Polte-Rudolf, C. (2015). An overview of the phenological observation network and the phenological database of Germany's national meteorological service (Deutscher Wetterdienst). *Advances in Science and Research, 11*, 93-99

Katal, N., Rzanny, M., Mader, P., & Waldchen, J. (2022). Deep Learning in Plant Phenological Research: A Systematic Literature Review. *Front Plant Sci, 13*, 805738

Kattenborn, T., Schiefer, F., Frey, J., Feilhauer, H., Mahecha, M.D., & Dormann, C.F. (2022). Spatially autocorrelated training and validation samples inflate performance assessment of convolutional neural networks. *ISPRS Open Journal of Photogrammetry and Remote Sensing, 5*, 100018

Kavats, O., Khramov, D., Sergieieva, K., & Vasyliev, V. (2019). Monitoring harvesting by time series of Sentinel-1 SAR data. *Remote Sensing, 11*, 2496

Ke, G., Meng, Q., Finley, T., Wang, T., Chen, W., Ma, W., Ye, Q., & Liu, T.-Y. (2017). LightGBM: A Highly Efficient Gradient Boosting Decision Tree. In, *Neural Information Processing Systems*

Keenan, T.F., Richardson, A.D., & Hufkens, K. (2020). On quantifying the apparent temperature sensitivity of plant phenology. *New Phytol, 225*, 1033-1040

Kooistra, L., Berger, K., Brede, B., Graf, L.V., Aasen, H., Roujean, J.-L., Machwitz, M., Schlerf, M., Atzberger, C., & Prikaziuk, E. (2023). Reviews and syntheses: Remotely sensed optical time series for monitoring vegetation productivity. *Biogeosciences Discussions, 2023*, 1-67

Kowalski, K., Senf, C., Hostert, P., & Pflugmacher, D. (2020). Characterizing spring phenology of temperate broadleaf forests using Landsat and Sentinel-2 time series. *International Journal of Applied Earth Observation and Geoinformation, 92*, 102172

Le Roux, R., Furusho-Percot, C., Deswarte, J.-C., Bancal, M.-O., Chenu, K., de Noblet-Ducoudré, N., de Cortázar-Atauri, I.G., Durand, A., Bulut, B., Maury, O., Décome, J., & Launay, M. (2024). Mapping the race between crop phenology and climate risks for wheat in France under climate change. *Scientific Reports, 14*, 8184

Lechner, A.M., Stein, A., Jones, S.D., & Ferwerda, J.G. (2009). Remote sensing of small and linear features: Quantifying the effects of patch size and length, grid position and detectability on land cover mapping. *Remote Sensing of Environment, 113*, 2194-2204

Li, H., Zhao, J.Y., Yan, B.Q., Yue, L.W., & Wang, L.C. (2022a). Global DEMs vary from one to another: an evaluation of newly released Copernicus, NASA and AW3D30 DEM on selected terrains of China using ICESat-2 altimetry data. *International Journal of Digital Earth, 15*, 1149-1168

Li, R.M., Xia, H.M., Zhao, X.Y., & Guo, Y. (2023). Mapping evergreen forests using new phenology index, time series Sentinel-1/2 and Google Earth Engine. *Ecological Indicators, 149*, 110157

Li, W.J., Xin, Q.C., Zhou, X.W., Zhang, Z.C., & Ruan, Y.J. (2021). Comparisons of numerical phenology models and machine learning methods on predicting the spring onset of natural vegetation across the Northern Hemisphere. *Ecological Indicators, 131*, 108126

Li, Z., Shen, H., Weng, Q., Zhang, Y., Dou, P., & Zhang, L. (2022b). Cloud and cloud shadow detection for optical satellite imagery: Features, algorithms, validation, and prospects. *ISPRS Journal of Photogrammetry and Remote Sensing, 188*, 89-108

Liu, G., Chuine, I., Denéchère, R., Jean, F., Dufrêne, E., Vincent, G., Berveiller, D., & Delpierre, N. (2021a). Higher sample sizes and observer inter-calibration are needed for reliable scoring of leaf phenology in trees. *Journal of Ecology, 109*, 2461-2474

Liu, H., Zhou, B., Bai, Z., Zhao, W., Zhu, M., Zheng, K., Yang, S., & Li, G. (2023). Applicability Assessment of Multi-Source DEM-Assisted InSAR Deformation Monitoring Considering Two Topographical Features. In, *Land*

Liu, L.C., Cao, R.Y., Chen, J., Shen, M.G., Wang, S., Zhou, J., & He, B.B. (2022). Detecting crop phenology from vegetation index time-series data by improved shape model fitting in each phenological stage. *Remote Sensing of Environment, 277*, 113060

Liu, Y., Qian, J., & Yue, H. (2021b). Comprehensive evaluation of Sentinel-2 red edge and shortwave-infrared bands to estimate soil moisture. *IEEE Journal of Selected Topics in Applied Earth Observations and Remote Sensing, 14*, 7448-7465

Lobert, F., Holtgrave, A.-K., Schwieder, M., Pause, M., Vogt, J., Gocht, A., & Erasmi, S. (2021). Mowing event detection in permanent grasslands: Systematic evaluation of input features from Sentinel-1, Sentinel-2, and Landsat 8 time series. *Remote Sensing of Environment, 267*, 112751

Lobert, F., Löw, J., Schwieder, M., Gocht, A., Schlund, M., Hostert, P., & Erasmi, S. (2023). A deep learning approach for deriving winter wheat phenology from optical and SAR time series at field level. *Remote Sensing of Environment, 298*, 113800

Lopez-Sanchez, J.M., Cloude, S.R., & Ballester-Berman, J.D. (2011). Rice phenology monitoring by means of SAR polarimetry at X-band. *IEEE Transactions on Geoscience and Remote Sensing, 50*, 2695-2709

Lopez-Sanchez, J.M., Vicente-Guijalba, F., Ballester-Berman, J.D., & Cloude, S.R. (2013). Polarimetric response of rice fields at C-band: Analysis and phenology retrieval. *IEEE Transactions on Geoscience and Remote Sensing, 52*, 2977-2993

Löw, J., Hill, S., Otte, I., Thiel, M., Ullmann, T., & Conrad, C. (2024). How Phenology Shapes Crop-Specific Sentinel-1 PolSAR Features and InSAR Coherence across Multiple Years and Orbits. *Remote Sensing, 16*, 2791

Löw, J., Ullmann, T., & Conrad, C. (2021). The Impact of Phenological Developments on Interferometric and Polarimetric Crop Signatures Derived from Sentinel-1: Examples from the DEMMIN Study Site (Germany). *Remote Sensing, 13*, 2951

Lu, X., Zhou, J., Yang, R., Yan, Z., Lin, Y., Jiao, J., & Liu, F. (2023). Automated Rice Phenology Stage Mapping Using UAV Images and Deep Learning. *Drones, 7*, 83

Ma, Y.Y., Shen, Y.L., Guan, H.X., Wang, J., & Hu, C.L. (2023). A novel approach to detect the spring corn phenology using layered strategy. *International Journal of Applied Earth Observation and Geoinformation, 122*, 103422

MacBean, N., Maignan, F., Peylin, P., Bacour, C., Bréon, F.M., & Ciais, P. (2015). Using satellite data to improve the leaf phenology of a global terrestrial biosphere model. *Biogeosciences, 12*, 7185-7208

Mandal, D., Ratha, D., Bhattacharya, A., Kumar, V., McNairn, H., Rao, Y.S., & Frery, A.C. (2020). A Radar Vegetation Index for Crop Monitoring Using Compact Polarimetric SAR Data. *IEEE Transactions on Geoscience and Remote Sensing, 58*, 6321-6335

Marini, L., Scotton, M., Klimek, S., Isselstein, J., & Pecile, A. (2007). Effects of local factors on plant species richness and composition of Alpine meadows. *Agriculture Ecosystems & Environment, 119*, 281-288

Marsh, C.B., Harder, P., & Pomeroy, J.W. (2023). Validation of FABDEM, a global bare-earth elevation model, against UAV-lidar derived elevation in a complex forested mountain catchment. *Environmental Research Communications, 5*, 031009

Maurya, A.K., Khangarot, L.S., & Singh, D. (2023). Crop Phenology Studies Using RGB Drone Data. In, *2023 International Conference on Electrical, Electronics, Communication and Computers (ELEXCOM)* (pp. 1-5)

McMaster, G.S., & Wilhelm, W.W. (1997). Growing degree-days: one equation, two interpretations. *Agricultural and Forest Meteorology, 87*, 291-300

Mei, X., Zhu, Q., Ma, L., Zhang, D., Wang, Y., & Hao, W. (2018). Effect of stand origin and slope position on infiltration pattern and preferential flow on a Loess hillslope. *Land Degradation & Development, 29*, 1353-1365

Mercier, A., Betbeder, J., Baudry, J., Le Roux, V., Spicher, F., Lacoux, J., Roger, D., & Hubert-Moy, L. (2020). Evaluation of Sentinel-1 & 2 time series for predicting wheat and rapeseed phenological stages. *ISPRS Journal of Photogrammetry and Remote Sensing, 163*, 231-256

Meroni, M., d'Andrimont, R., Vrieling, A., Fasbender, D., Lemoine, G., Rembold, F., Seguini, L., & Verhegghen, A. (2021). Comparing land surface phenology of major European crops as derived from SAR and multispectral data of Sentinel-1 and -2. *Remote Sens Environ, 253*, 112232

Mimić, G., Mishra, A.K., Marković, M., Živaljević, B., Pavlović, D., & Marko, O. (2025). Machine Learning-Based Harvest Date Detection and Prediction Using SAR Data for the Vojvodina Region (Serbia). In, *Sensors*

Morellato, L.P.C., Alberton, B., Alvarado, S.T., Borges, B., Buisson, E., Camargo, M.G.G., Cancian, L.F., Carstensen, D.W., Escobar, D.F.E., Leite, P.T.P., Mendoza, I., Rocha, N.M.W.B., Soares, N.C., Silva, T.S.F., Staggemeier, V.G., Streher, A.S., Vargas, B.C., & Peres, C.A. (2016). Linking plant phenology to conservation biology. *Biological Conservation, 195*, 60-72

Mullissa, A., Vollrath, A., Odongo-Braun, C., Slagter, B., Balling, J., Gou, Y., Gorelick, N., & Reiche, J. (2021). Sentinel-1 SAR Backscatter Analysis Ready Data Preparation in Google Earth Engine. *Remote Sensing, 13*, 1954

Nieto, L., Schwalbert, R., Prasad, P.V.V., Olson, B.J.S.C., & Ciampitti, I.A. (2021). An integrated approach of field, weather, and satellite data for monitoring maize phenology. *Scientific Reports, 11*, 15711

Nietupski, T.C., Kennedy, R.E., Temesgen, H., & Kerns, B.K. (2021). Spatiotemporal image fusion in Google Earth Engine for annual estimates of land surface phenology in a heterogenous landscape. *International Journal of Applied Earth Observation and Geoinformation, 99*, 102323

Olson, D., Chatterjee, A., & Franzen, D.W. (2019). Can we select sugarbeet harvesting dates using drone-based vegetation indices? *Agronomy Journal, 111*, 2619-2624

Patel, D., & Franklin, K.A. (2009). Temperature-regulation of plant architecture. *Plant Signal Behav, 4*, 577-579

Pei, J., Tan, S., Zou, Y., Liao, C., He, Y., Wang, J., Huang, H., Wang, T., Tian, H., Fang, H., Wang, L., & Huang, J. (2025). The role of phenology in crop yield prediction: Comparison of ground-based phenology and remotely sensed phenology. *Agricultural and Forest Meteorology, 361*, 110340

Pipia, L., Munoz-Mari, J., Amin, E., Belda, S., Camps-Valls, G., & Verrelst, J. (2019). Fusing optical and SAR time series for LAI gap fillingwith multioutput Gaussian processes. *Remote Sens Environ, 235*

Povey, A.C., & Grainger, R.G. (2015). Known and unknown unknowns: uncertainty estimation in satellite remote sensing. *Atmos. Meas. Tech., 8*, 4699-4718

Revill, A., Florence, A., MacArthur, A., Hoad, S.P., Rees, R.M., & Williams, M. (2019). The Value of Sentinel-2 Spectral Bands for the Assessment of Winter Wheat Growth and Development. *Remote Sensing, 11*, 2050

Rezaei, E.E., Siebert, S., & Ewert, F. (2017). Climate and management interaction cause diverse crop phenology trends. *Agricultural and Forest Meteorology, 233*, 55-70

Rezaei, E.E., Siebert, S., Hüging, H., & Ewert, F. (2018). Climate change effect on wheat phenology depends on cultivar change. *Scientific Reports, 8*, 4891

Schlund, M., & Erasmi, S. (2020). Sentinel-1 time series data for monitoring the phenology of winter wheat. *Remote Sensing of Environment, 246*, 111814

Senaras, C., Grady, M., Rana, A.S., Nieto, L., Ciampitti, I., Holden, P., Davis, T., & Wania, A. (2024). Detection of Maize Crop Phenology Using Planet Fusion. *Remote Sensing, 16*, 2730

Shang, J., Liu, J., Poncos, V., Geng, X., Qian, B., Chen, Q., Dong, T., Macdonald, D., Martin, T., & Kovacs, J. (2020). Detection of crop seeding and harvest through analysis of time-series Sentinel-1 interferometric SAR data. *Remote Sensing, 12*, 1551

Shao, Y., Lunetta, R.S., Wheeler, B., Iiames, J.S., & Campbell, J.B. (2016). An evaluation of time-series smoothing algorithms for land-cover classifications using MODIS-NDVI multi-temporal data. *Remote Sensing of Environment, 174*, 258-265

Shimada, M., Itoh, T., Motooka, T., Watanabe, M., Shiraishi, T., Thapa, R., & Lucas, R. (2014). New global forest/non-forest maps from ALOS PALSAR data (2007–2010). *Remote Sensing of Environment, 155*, 13-31

Shrestha, A., Bheemanahalli, R., Adeli, A., Samiappan, S., Czarnecki, J.M.P., McCraine, C.D., Reddy, K.R., & Moorhead, R. (2023). Phenological stage and vegetation index for predicting corn yield under rainfed environments. *Frontiers in Plant Science, 14*, 1168732

Singh, S. (2018). Understanding the role of slope aspect in shaping the vegetation attributes and soil properties in Montane ecosystems. *Tropical Ecology, 59*, 417-430

Sitokonstantinou, V., Koukos, A., Tsoumas, I., Bartsotas, N.S., Kontoes, C., & Karathanassi, V. (2023). Fuzzy clustering for the within-season estimation of cotton phenology. *PLoS ONE, 18*, e0282364

Tedesco, D., de Oliveira, M.F., dos Santos, A.F., Silva, E.H.C., Rolim, G.D., & da Silva, R.P. (2021). Use of remote sensing to characterize the phenological development and to predict sweet potato yield in two growing seasons. *European Journal of Agronomy, 129*, 126337

Templ, B., Koch, E., Bolmgren, K., Ungersböck, M., Paul, A., Scheifinger, H., Rutishauser, T., Busto, M., Chmielewski, F.-M., & Hájková, L. (2018). Pan European Phenological database (PEP725): a single point of access for European data. *International Journal of Biometeorology, 62*, 1109-1113

Tian, F., Cai, Z.Z., Jin, H.X., Hufkens, K., Scheifinger, H., Tagesson, T., Smets, B., Van Hoolst, R., Bonte, K., Ivits, E., Tong, X.Y., Ardö, J., & Eklundh, L. (2021). Calibrating vegetation phenology from Sentinel-2 using eddy covariance, PhenoCam, and PEP725 networks across Europe. *Remote Sensing of Environment, 260*, 112456

Tomczak, M. (1998). Spatial interpolation and its uncertainty using automated anisotropic inverse distance weighting (IDW)-cross-validation/jackknife approach. *Journal of Geographic Information and Decision Analysis, 2*, 18-30

Tran, K.H., Zhang, X., Ye, Y., Shen, Y., Gao, S., Liu, Y., & Richardson, A. (2023). HP-LSP: A reference of land surface phenology from fused Harmonized Landsat and Sentinel-2 with PhenoCam data. *Sci Data, 10*, 691

Tran, K.H., Zhang, X., Zhang, H.K., Shen, Y., Ye, Y., Liu, Y., Gao, S., & An, S. (2025). A transformer-based model for detecting land surface phenology from the irregular harmonized Landsat and Sentinel-2 time series across the United States. *Remote Sensing of Environment, 320*, 114656

Tucker, C.J. (1978). A comparison of satellite sensor bands for vegetation monitoring. *Photogrammetric Engineering and Remote Sensing, 44*, 1369-1380

Veloso, A., Mermoz, S., Bouvet, A., Le Toan, T., Planells, M., Dejoux, J.-F., & Ceschia, E. (2017). Understanding the temporal behavior of crops using Sentinel-1 and Sentinel-2-like data for agricultural applications. *Remote Sensing of Environment, 199*, 415-426

Vijaywargiya, J., & Nidamanuri, R.R. (2023). Crop Phenology Extraction Using Big Geospatial Datacube, 1-4

Vina, A., Gitelson, A.A., Rundquist, D.C., Keydan, G., Leavitt, B., & Schepers, J. (2004). Monitoring maize (Zea mays L.) phenology with remote sensing. *Agronomy Journal, 96*, 1139-1147

Viña, A., Liu, W., Zhou, S., Huang, J., & Liu, J. (2016). Land surface phenology as an indicator of biodiversity patterns. *Ecological Indicators, 64*, 281-288

Viswanathan, M., Scheidegger, A., Streck, T., Gayler, S., & Weber, T.K.D. (2022). Bayesian multi-level calibration of a process-based maize phenology model. *Ecological Modelling, 474*, 110154

Wadoux, A.M.-C., Heuvelink, G.B., De Bruin, S., & Brus, D.J. (2021). Spatial cross-validation is not the right way to evaluate map accuracy. *Ecological Modelling, 457*, 109692

Wang, H., Magagi, R., Goïta, K., Trudel, M., McNairn, H., & Powers, J. (2019a). Crop phenology retrieval via polarimetric SAR decomposition and Random Forest algorithm. *Remote Sensing of Environment, 231*, 111234

Wang, H.Q., Magagi, R., Goïta, K., Trudel, M., McNairn, H., & Powers, J. (2019b). Crop phenology retrieval via polarimetric SAR decomposition and Random Forest algorithm. *Remote Sensing of Environment, 231*, 111234

Wang, J., Song, G.Q., Liddell, M., Morellato, P., Lee, C.K.F., Yang, D.D., Alberton, B., Detto, M., Ma, X.L., Zhao, Y.Y., Yeung, H.C.H., Zhang, H.S., Ng, M., Nelson, B.W., Huete, A., & Wu, J. (2023). An ecologically-constrained deep learning model for tropical leaf phenology monitoring using PlanetScope satellites. *Remote Sensing of Environment, 286*, 113429

Worrall, G., Judge, J., Boote, K., & Rangarajan, A. (2023). In-season crop phenology using remote sensing and model-guided machine learning. *Agronomy Journal, 115*, 1214-1236

Wu, B., Tian, F., Nabil, M., Bofana, J., Lu, Y., Elnashar, A., Beyene, A.N., Zhang, M., Zeng, H., & Zhu, W. (2023). Mapping global maximum irrigation extent at 30m resolution using the irrigation performances under drought stress. *Global Environmental Change, 79*, 102652

Xia, J., Niu, S., Ciais, P., Janssens, I.A., Chen, J., Ammann, C., Arain, A., Blanken, P.D., Cescatti, A., Bonal, D., Buchmann, N., Curtis, P.S., Chen, S., Dong, J., Flanagan, L.B., Frankenberg, C., Georgiadis, T., Gough, C.M., Hui, D., Kiely, G., Li, J., Lund, M., Magliulo, V., Marcolla, B., Merbold, L., Montagnani, L., Moors, E.J., Olesen, J.E., Piao, S., Raschi, A., Roupsard, O., Suyker, A.E., Urbaniak, M., Vaccari, F.P., Varlagin, A., Vesala, T., Wilkinson, M., Weng, E., Wohlfahrt, G., Yan, L., & Luo, Y. (2015). Joint control of terrestrial gross primary productivity by plant phenology and physiology. *Proc Natl Acad Sci U S A, 112*, 2788-2793

Xin, Q.C., Li, J., Li, Z.M., Li, Y.M., & Zhou, X.W. (2020). Evaluations and comparisons of rule-based and machine-learning-based methods to retrieve satellite-based vegetation phenology using MODIS and USA National Phenology Network data. *International Journal of Applied Earth Observation and Geoinformation, 93*, 102189

Yang, H.J., Pan, B., Li, N., Wang, W., Zhang, J., & Zhang, X.L. (2021). A systematic method for spatio-temporal phenology estimation of paddy rice using time series Sentinel-1 images. *Remote Sensing of Environment, 259*, 112394

Yang, J., Shi, H., Xie, Q., Lopez-Sanchez, J.M., Peng, X., Yu, J., & Chen, L. (2024a). Crop Phenology Estimation in Rice Fields Using Sentinel-1 GRD SAR Data and Machine Learning-Aided Particle Filtering Approach. *Int. Arch. Photogramm. Remote Sens. Spatial Inf. Sci., XLVIII-1-2024*, 799-804

Yang, J.L., Dong, J.W., Liu, L., Zhao, M.M., Zhang, X.Y., Li, X.C., Dai, J.H., Wang, H.J., Wu, C.Y., You, N.S., Fang, S.B., Pang, Y., He, Y.L., Zhao, G.S., Xiao, X.M., & Ge, Q.S. (2023a). A robust and unified land surface phenology algorithm for diverse biomes and growth cycles in China by using harmonized Landsat and Sentinel-2 imagery. *ISPRS Journal of Photogrammetry and Remote Sensing, 202*, 610-636

Yang, K., Liu, L., & Wen, Y. (2024b). The impact of Bayesian optimization on feature selection. *Scientific Reports, 14*, 3948

Yang, Q., Shi, L., Han, J., Chen, Z., & Yu, J. (2022). A VI-based phenology adaptation approach for rice crop monitoring using UAV multispectral images. *Field Crops Research, 277*, 108419

Yang, Z.J., Diao, C.Y., & Gao, F. (2023b). Towards Scalable Within-Season Crop Mapping With Phenology Normalization and Deep Learning. *IEEE Journal of Selected Topics in Applied Earth Observations and Remote Sensing, 16*, 1390-1402

Yeasin, M., Haldar, D., Kumar, S., Paul, R.K., & Ghosh, S. (2022). Machine Learning Techniques for Phenology Assessment of Sugarcane Using Conjunctive SAR and Optical Data. *Remote Sensing, 14*, 3249

Yue, J., Yihan, Y., Jianing, S., Ting, L., Nianxu, X., Haikuan, F., Yihao, W., Xin, X., Yinghao, L., Wei, G., Yuanyuan, F., Hongbo, Q., Xinming, M., & and Wang, J. (2025). Winter wheat harvest detection via Sentinel-2 MSI images. *International Journal of Remote Sensing, 46*, 2482-2500

Zamani-Noor, N., & Feistkorn, D. (2022). Monitoring Growth Status of Winter Oilseed Rape by NDVI and NDYI Derived from UAV-Based Red–Green–Blue Imagery. *Agronomy, 12*, 2212

Zanaga, D., Van De Kerchove, R., De Keersmaecker, W., Souverijns, N., Brockmann, C., Quast, R., Wevers, J., Grosu, A., Paccini, A., Vergnaud, S., Cartus, O., Santoro, M., Fritz, S., Georgieva, I., Lesiv, M., Carter, S., Herold, M., Li, L., Tsendbazar, N.-E., Ramoino, F., & Arino, O. (2021). ESA WorldCover 10 m 2020 v100. In: Zenodo

Zeng, L., Wardlow, B.D., Xiang, D., Hu, S., & Li, D. (2020). A review of vegetation phenological metrics extraction using time-series, multispectral satellite data. *Remote Sensing of Environment, 237*, 111511

Zhang, L.L., Zhang, Z., Luo, Y.C., Cao, J., Xie, R.Z., & Li, S.K. (2021). Integrating satellite-derived climatic and vegetation indices to predict smallholder maize yield using deep learning. *Agricultural and Forest Meteorology, 311*, 108666

Zhao, W.Z., Qu, Y., Zhang, L.Q., & Li, K.Y. (2022). Spatial-aware SAR-optical time-series deep integration for crop phenology tracking. *Remote Sensing of Environment, 276*, 113046

Zheng, Z., Zhu, W., Chen, G., Jiang, N., Fan, D., & Zhang, D. (2016). Continuous but diverse advancement of spring-summer phenology in response to climate warming across the Qinghai-Tibetan Plateau. *Agricultural and Forest Meteorology, 223*, 194-202

Zhou, J., Jia, L., Menenti, M., & Gorte, B. (2016). On the performance of remote sensing time series reconstruction methods – A spatial comparison. *Remote Sensing of Environment, 187*, 367-384

Zhou, Q., Guan, K., Wang, S., Hipple, J., & Chen, Z. (2024). From satellite-based phenological metrics to crop planting dates: Deriving field-level planting dates for corn and soybean in the U.S. Midwest. *ISPRS Journal of Photogrammetry and Remote Sensing, 216*, 259-273

Zhou, X., Xin, Q., Dai, Y., Li, W., & Qiao, H. (2021). A deep-learning-based experiment for benchmarking the performance of global terrestrial vegetation phenology models. *Global Ecology and Biogeography, 30*, 2178-2199

Ziegler, K., Pollinger, F., Böll, S., & Paeth, H. (2020). Statistical modeling of phenology in Bavaria based on past and future meteorological information. *Theoretical and Applied Climatology, 140*, 1467-1481

| Figure # | Caption |
| --- | --- |
| Figure 1 | Distribution of selected DWD stations on a digital elevation model (DEM; A) with a 5-km square buffers around three example stations (B-D; white patches indicate non-target areas) showing the Crop Type Map (CTM). |
| Figure 2 | Time series of RS and climate parameters for winter wheat at different BBCH stages in 2020 growing season at Hohenbachen in Bavaria; G: Green band, SWIR1: Short Wave InfraRed 1, NDYI: Normalized Difference Infrared Index, VARI: Visible Atmospherically Resistant Index, VH: Vertical transmit and Horizontal receiver, VV: Vertical transmit and Vertical receiver (VH; dB) |
| Figure 3 | Flowchart of the proposed framework. All abbreviations are defined in the text. |
| Figure 4 | Selected features for each crop based on feature optimization. |
| Figure 5 | Scatter plot of predictions of unseen data for each crop and BBCH. The dashed black lines are the deviation of ±15 days from the 1:1 line. The red lines are the linear regressions between observations and prediction for each BBCH stage. |
| Figure 6 | Bar plot of the MAE (on left vertical axis) and line plot of $R^2$ (on right vertical axis) for each crop in each BBCH. The dashed red line is a deviation of ±6 days from the prefect prediction. The percentage shows the proportion of the prediction that has differences with observation within ±6 days. |
| Figure 7 | Spatial difference between estimated and observed phenology for all crops at seeding stage across Germany. The color shows the number of days between estimated and observed values, and the numbers represent the concordant percentage of data in each range. |
| Figure 8 | Spatial difference between estimated and observed phenology for all crops at harvesting stage across Germany. The color shows the difference amount between estimated and observed values and numbers in each color represent the percentage of data in each range. |
| Figure 9 | Differences between MAE (days) of model predictions for each year and each crop in whole stages together. All in the year means for whole years together. |
| Figure 10 | Density plot of predictions versus observations for winter wheat in different phenology stage. The dashed black lines are the deviation of ±6 days from the prefect prediction. |
| Figure A1 | Distribution of DWD and climate stations across Germany. |
| Figure A2 | Red pixels show areas that are classified as a crop type while they are classified as forest (left) in ESA Land Use Land Cover (LULC) and (right) in JAXA Forest/Non-Forest (FNF) classification. The table at the bottom represents the statistics of difference between (in meters) CDEM and FABDEM in all pixels for each crop type in all 862 DWD stations. Comparison of CTM data with ESA World Cover (Zanaga et al. 2021) and JAXA forest/non-forest data (Shimada et al. 2014) for 2020, assessing misclassified crop boundaries and differences between FABDEM and CDEM at 862 DWD stations for eight crops. Results support the use of FABDEM for spatial phenology estimation in agroforestry regions like Germany. |
| Figure A3 | The difference between MAE of the standardized feature sets and best feature set for each crop and each BBCH. |
| Figure A4 | The difference between $R^2$ of the standardized feature sets and removing the static features for winter wheat. |
| Figure A5 | The difference between MAE of the standardized feature sets and removing the static features for winter wheat. |
| Figure A6 | The difference between MAE of the standardized feature sets and removing the static features for winter wheat. |
| Figure A7 | The difference between MAE of all stations (including low-precision ones) and high-precision geolocation stations for winter wheat. |
| Figure A8 | The difference between MAE of all stations (including low-precision ones) and high-precision geolocation stations for winter wheat. |
| Figure A9 | Density plot of predictions versus observations for maize in different phenology stage. The dashed black lines are the deviation of ±6 days from the prefect prediction. |
| Figure A10 | Density plot of predictions versus observations for spring barley in different phenology stage. The dashed black lines are the deviation of ±6 days from the prefect prediction. |
| Figure A11 | Density plot of predictions versus observations for spring oat in different phenology stage. The dashed black lines are the deviation of ±6 days from the prefect prediction. |
| Figure A12 | Density plot of predictions versus observations for sugar beet in different phenology stage. The dashed black lines are the deviation of ±6 days from the prefect prediction. |
| Figure A13 | Density plot of predictions versus observations for winter barley in different phenology stage. The dashed black lines are the deviation of ±6 days from the prefect prediction. |
| Figure A14 | Density plot of predictions versus observations for winter rapeseed in different phenology stage. The dashed black lines are the deviation of ±6 days from the prefect prediction. |
| Figure A15 | Density plot of predictions versus observations for winter rye in different phenology stage. The dashed black lines are the deviation of ±6 days from the prefect prediction. |